\documentclass[10pt]{article}
\usepackage{amsmath,amsthm,verbatim,amssymb,amsfonts,amscd, graphicx}
\usepackage{graphics}
\usepackage{centernot}
\theoremstyle{plain}
\newtheorem{theorem}{Theorem}

\newtheorem{lemma}{Lemma}
\newtheorem*{remark}{Remark}
\newtheorem{proposition}{Proposition}

\theoremstyle{definition}
\newtheorem{definition}{Definition}
\usepackage{bbm}
\usepackage[colorlinks,linkcolor=blue,citecolor=blue]{hyperref}

\usepackage[bottom]{footmisc}
\usepackage{caption}
\usepackage{subcaption}
\usepackage{helvet}  
\usepackage{courier}  
\usepackage{url}  
\usepackage{graphicx}  
\usepackage{multirow}
\usepackage{amsthm}
\usepackage{color}
\usepackage{MnSymbol}
\usepackage{makecell}
\usepackage{arydshln}
\usepackage{amsmath}
\usepackage{xcolor}
\usepackage{caption} 
\usepackage{natbib}
\usepackage{textcomp}
\usepackage{wrapfig}
\usepackage{algorithm}
\usepackage{algorithmic}

\usepackage{csquotes}

\newcommand{\E}{\mathbb{E}}

\begin{document}
\title{Theoretically Motivated Data Augmentation and Regularization for Portfolio Construction}
\author{Liu Ziyin$_1$, Kentaro Minami$_2$, Kentaro Imajo$_2$\\
\textit{$_1$Department of Physics, The University of Tokyo}\\
\textit{$_2$Preferred Networks, Inc.}}
\maketitle

\begin{abstract}
    The task we consider is portfolio construction in a speculative market, a fundamental problem in modern finance. While various empirical works now exist to explore deep learning in finance, the theory side is almost non-existent. In this work, we focus on developing a theoretical framework for understanding the use of data augmentation for deep-learning-based approaches to quantitative finance. The proposed theory clarifies the role and necessity of data augmentation for finance; moreover, our theory implies that a simple algorithm of injecting a random noise of strength $\sqrt{|r_{t-1}|}$ to the observed return $r_{t}$ is better than not injecting any noise and a few other financially irrelevant data augmentation techniques.\footnote{This is the full-length version of our work published at the 3rd ACM International Conference on AI in Finance (ICAIF'22). See \url{https://doi.org/10.1145/3533271.3561720} for the shorter published version.\\ The code is available at: \url{https://github.com/pfnet-research/Finance_data_augmentation_ICAIF2022} }
\end{abstract}

\section{Introduction}
There is an increasing interest in applying machine learning methods to problems in the finance industry. This trend has been expected for almost forty years \citep{fama1970efficient}, when well-documented and fine-grained (minute-level) data of stock market prices became available. In fact, the essence of modern finance is fast and accurate large-scale data analysis \citep{goodhart1997high}, and it is hard to imagine that machine learning should not play an increasingly crucial role in this field. In contemporary research, the central theme in machine-learning based finance is to apply existing deep learning models to financial time-series prediction problems \citep{imajo2020deep, buehler2019deep, jay2020stochastic, imaki2021no, jiang2017deep, fons2020evaluating, lim2019enhancing, Zhang_2020}, which have demonstrated the hypothesized usefulness of deep learning for the financial industry.

However, one major existing gap in this interdisciplinary field of deep-learning finance is the lack of a theory relevant to justify finance-oriented algorithm design. The goal of this work is to propose such a framework, where machine learning practices are analyzed in a traditional financial-economic utility theory setting. Our theory implies that a simple theoretically motivated data augmentation technique is better for the task portfolio construction than not injecting any noise at all or some naive noise injection methods that have no theoretical justification. To summarize, our main contributions are (1) to demonstrate how we can use utility theory to analyze practices of deep-learning-based finance, and (2) to theoretically study the role of data augmentation in the deep-learning-based portfolio construction problem. \textbf{Organization}: the next section discusses the main related works; Section 3 provides the requisite finance background for understanding this work; Section 4 presents our theoretical contributions, which is a framework for understanding machine-learning practices in the portfolio construction problem; Section 5 describes how to practically implement the theoretically motivated algorithm; section 6 validates the theory with experiments on toy and real data. 

\vspace{-2mm}
\section{Related Works}
\vspace{-0mm}

\textbf{Existing deep learning finance methods}. In recent years, various empirical approaches to apply state-of-the-art deep learning methods to finance have been proposed \citep{imajo2020deep, ito2020tradercompany, buehler2019deep, jay2020stochastic, imaki2021no, jiang2017deep, fons2020evaluating}. The interested readers are referred to \citep{ozbayoglu2020deep} for  detailed descriptions of existing works. However, we notice that one crucial gap is the complete lack of theoretical analysis or motivation in this interdisciplinary field of AI-finance. This work makes one initial step to bridge this gap. 
One theme of this work is that finance-oriented prior knowledge and inductive bias is required to design the relevant algorithms. For example, \citet{ziyin2020neural} shows that incorporating prior knowledge into architecture design is key to the success of neural networks and applied neural networks with periodic activation functions to the problem of financial index prediction. \citet{imaki2021no} shows how to incorporate no-transaction prior knowledge into network architecture design when transaction cost is incorporated. 

\begin{figure}
\vspace{0em}
\centering
    \includegraphics[width=0.4\linewidth]{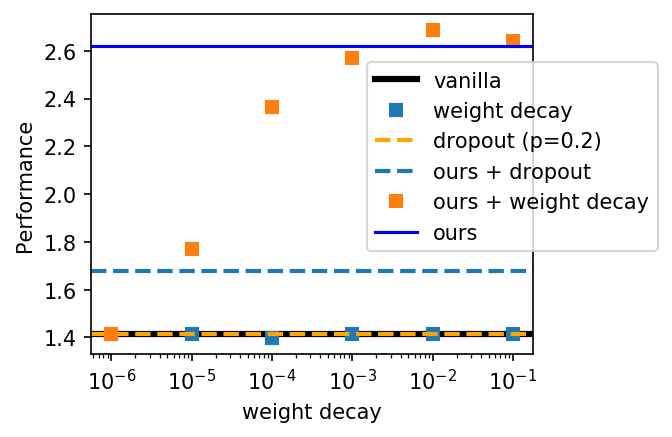}
    \vspace{-1em}
    \caption{\small Performance (measured by the Sharpe ratio) of various algorithms on MSFT (Microsoft) from 2018-2020. Directly applying generic machine learning methods, such as weight decay, fails to improve the vanilla model. The proposed method show significant improvement. }\label{fig: motivation}
    \vspace{-1em}
\end{figure}

In fact, most generic and popular machine learning techniques are proposed and have been tested for standard ML tasks such as image classification or language processing. Directly applying the ML methods that work for image tasks is unlikely to work well for financial tasks, where the nature of the data is different. See Figure~\ref{fig: motivation}, where we show the performance of a neural network directly trained to maximize wealth return on MSFT during 2019-2020. Using popular, generic deep learning techniques such as weight decay or dropout does not result in any improvement over the baseline. In contrast, our theoretically motivated method does. Combining the proposed method with weight decay has the potential to improve the performance a little further, but the improvement is much lesser than the improvement of using the proposed method over the baseline. This implies that a generic machine learning method is unlikely to capture well the inductive biases required to tackle a financial task. The present work proposes to fill this gap by showing how finance knowledge can be incorporated into algorithm design.

\textbf{Data augmentation}. Consider a training loss function of the additive form $L = \frac{1}{N} \sum_i \ell(x_i, y_i)$ for $N$ pairs of training data points $\{(x_i, y_i)\}_{i=1}^N$. Data augmentation amounts to defining an underlying data-dependent distribution and generating new data points stochastically from this underlying distribution. A general way to define data augmentation is to start with a datum-level training loss and transform it to an expectation over an augmentation distribution $P\left(z |(x_i, y_i) \right)$ \citep{dao2019kernel}, 
$\ell(x_i, y_i) \to \mathbb{E}_{(z_i, g_i) \sim P(z, g | (x_i, y_i))} [\ell(z_i, g_i) ]$,
and the total training loss function becomes
\begin{equation}
    L_{\rm aug} = \frac{1}{N} \sum_{i=1}^N \mathbb{E}_{(z_i, g_i) \sim P(z, g | (x_i, y_i))} [\ell(z_i, g_i) ].
\end{equation}
One common example of data augmentation is injecting isotropic gaussian noise to the input \citep{shorten2019survey, fons2020evaluating}, which is equivalent to setting $P(z, g | (x_i, y_i)) \sim \delta(g - y_i)\exp\left[-(z - x_i)^{\rm T}(z- x_i)/(2\sigma^2) \right]$ for some specified strength $\sigma^2$. Despite the ubiquity of data augmentation in deep learning, existing works are often empirical in nature \citep{fons2020evaluating, zhong2020random, shorten2019survey, antoniou2017data}. For a relevant example, \citet{fons2020evaluating} empirically evaluates the effect of different types of data augmentation in a financial series prediction task. \citet{dao2019kernel} is one major recent theoretical work that tries to understand modern data augmentation theoretically; it shows that data augmentation is approximately learning in a special kernel. \citet{he2019data} argues that data augmentation can be seen as an effective regularization. However, no theoretically motivated data augmentation method for finance exists yet. One major challenge and achievement of this work is to develop a theory that bridges the traditional finance theory and machine learning methods. In the next section, we introduce the portfolio theory.

\vspace{0mm}
\section{Background: Markowitz Portfolio Theory}
\vspace{0mm}

How to make optimal investment in a financial market is the central concern of portfolio theory. One unfamiliar with the portfolio theory may easily confuse the task the portfolio construction with wealth maximization trading or future price prediction. Before we introduce the portfolio theory, we first stress that the task of portfolio construction is not equivalent to wealth maximization or accurate price prediction. One can construct an optimal portfolio without predicting the price or maximizing the wealth increase.

Consider a market with an equity (a stock) and a fixed-interest rate bond (a government bond). We denote the price of the equity at time step $t$ as $S_t$, and the \textit{price return} is defined as $r_t = \frac{S_{t+1} - S_{t}}{S_{t}}$, which is a random variable with variance $C_t$, and the expected return $g_t := \mathbb{E}[r_t]$. Our wealth at time step $t$ is  
    $W_t = M_t +  n_{t} S_{t}$, 
where $M_t$ is the amount of cash we hold, and $n_i$ the shares of stock we hold for the $i$-th stock. As in the standard finance literature, we assume that the shares are infinitely divisible. Usually, a positive $n$ denotes holding (long) and a negative $n$ denotes borrowing (short). The wealth we hold initially is $W_0 > 0$, and we would like to invest our money on the equity. We denote the relative value of the stock we hold as $\pi_{t} = \frac{n_{t} S_{t}}{W_t}$. $\pi$ is called a \textit{portfolio}. The central challenge in portfolio theory is to find the best $\pi$. At time $t$, our wealth is $W_t$; after one time step, our wealth changes due to a change in the price of the stock (setting the interest rate to be $0$): $\Delta W_{t} := W_{t+1} - W_{t} = W_t \pi_t r_t$. The goal is to maximize the wealth return $G_t := \pi_t \cdot r_t$ at every time step \textit{while minimizing risk}\footnote{It is important to not to confuse the \textit{price return} $r_t$ with the wealth return $G_t$.}. The risk is defined as the variance of the wealth change:
\begin{equation}
    R_t := R(\pi_t) :=  \text{Var}_{r_t}[G_t] = \left(\E[r_t^2] -g_t^2 \right)\pi_t^2 = \pi_t^2 C_t.
\end{equation}
The standard way to control risk is to introduce a ``risk regularizer" that punishes the portfolios with a large risk \citep{markowitz1959portfolio, rubinstein2002markowitz}.\footnote{In principle, any concave function in $G_t$ can be a risk regularizer from classical economic theory \citep{von1947theory}. One common alternative would be $R(G) = \log(G)$ \citep{kelly2011new}, and our framework can be easily extended to such cases.} Introducing a parameter $\lambda$ for the strength of regularization (the factor of $1/2$ appears for convention), we can now write down our objective:
\begin{equation}\label{eq: portfolio objective}
    \pi_t^* = \arg\max_\pi U(\pi) :=  \arg\max_\pi \left[\pi^{\rm T} G_t - \frac{\lambda}{2} R(\pi)  \right].
\end{equation}
Here, $U$ stands for the utility function; $\lambda$ can be set to be the desired level of risk-aversion. When $g_t$ and $C_t$ is known, this problem can be explicitly solved. However, one main problem in finance is that its data is highly limited and we only observe one particular \textit{realized} data trajectory, and $g_t$ and $C_t$ are hard to estimate. This fact motivates for the necessity of data augmentation and synthetic data generation in finance \citep{assefa2020generating}. In this paper, we treat the case where there is only one asset to trade in the market, and the task of utility maximization amounts to finding the best balance between cash-holding and investment. The equity we are treating is allowed to be a weighted combination of multiple stocks (a portfolio of some public fund manager, for example), and so our formalism is not limited to single-stock situations. In section~\ref{app sec: classical portfolio solution}, we discuss portfolio theory with multiple stocks.

\vspace{0mm}
\section{Portfolio Construction as a Training Objective}
\vspace{0mm}
Recent advances have shown that the financial objectives can be interpreted as training losses for an appropriately inserted neural-network model \citep{ziyin2019deep, buehler2019deep}. It should come as no surprise that the utility function \eqref{eq: portfolio objective} can be interpreted as a loss function. When the goal is portfolio construction, we parametrize the portfolio $\pi_t=\pi_\mathbf{w}(x_t)$ by a neural network with weights $\mathbf{w}$, and the utility maximization problem becomes a maximization problem over the weights of the neural network. The time-dependence is modeled through the input to the network $x_t$, which possibly consists of the available information at time $t$ for determining the future price.\footnote{It is helpful to imagine $x_t$ as, for example, the prices of the stocks in the past $10$ days.} The objective function (to be maximized) plus a pre-specified data augmentation transform $x_t\to z_t$ with underlying distribution $p(z|x_t)$ is then

{\small
\begin{equation}\label{eq: train objective}
    \pi^*_t = \arg\max_{\mathbf{w}} \left\{ \frac{1}{T} \sum_{t=1}^{T} \mathbb{E}_{t} \left[G_t(\pi_{\mathbf{w}}(z_t)) \right] - \lambda \text{Var}_t[ G_t(\pi_{\mathbf{w}}(z_t))] \right\},
\end{equation}\par}
where $\mathbb{E}_t := \mathbb{E}_{z_t \sim p(z|x_t)}$. In this work, we abstract away the details of the neural network to approximate $\pi$. We instead focus on studying the maximizers of this equation, which is a suitable choice when the underlying model is a neural network because one primary motivation for using neural networks in finance is that they are universal approximators and are often expected to find such maximizers \citep{buehler2019deep, imaki2021no}.   

The ultimate financial goal is to construct $\pi^*$ such that the utility function is maximized with respect to the true underlying distribution of $S_t$, which can be used as the generalization loss (to be maximized):
\begin{equation}\label{eq: test objective}
    \pi^*_t = \arg\max_{\pi_t}  \left\{ \mathbb{E}_{S_t} \left[ G_t(\pi) \right] - \lambda \text{Var}_{S_t}[ G_t(\pi)] \right\}.
\end{equation}
Note the difference in taking the expectation between Eq~\eqref{eq: train objective} and \eqref{eq: test objective} is that $\mathbb{E}_t$ is computed with respect to the training set we hold, while $\mathbb{E}_{S_t} := \mathbb{E}_{S_t\sim p(S_t)}$ is computed with respect to the underlying distribution of $S_t$ given its previous prices. We used the same short-hands for $\text{Var}_t$ and $\text{Var}_{S_t}$. Technically, the true utility we defined is an \textit{in-sample counterfactual} objective, which roughly evaluates the expected utility to be obtained if we restart from yesterday, which is a relevant measure for financial decision making. In Section~\ref{sec: stationary portfolio}, we also analyze the out-of-sample performance when the portfolio is static. 

\vspace{0mm}
\subsection{Standard Models of Stock Prices}
\vspace{0mm}
The expectations in the true objective Equation \eqref{eq: test objective} need to be taken with respect to the true underlying distribution of the stock price generation process. In general, the price follows the following stochastic process 
    $\Delta S_{t} = f(\{S_i\}_{i=1}^{t}) + g(\{S_i\}_{i=1}^{t})\eta_t$
for a zero-mean and unit variance random noise $\eta_t$; the term $f$ reflects the short-term predictability of the stock price based on past prices, and $g$ reflects the extent of unpredictability in the price. A key observation in finance is that $g$ is non-stationary (heteroskedastic) and price-dependent (multiplicative). One model is the geometric Brownian motion (GBM)
\begin{equation} \label{eq: geometric brownian motion}
    S_{t+1} = (1 +r) S_t + \sigma_t S_t \eta_t,
\end{equation} 
which is taken as the minimal standard model of the motion of stock prices \citep{mandelbrot1997variation, black1973pricing}; this paper also assumes the GBM model as the underlying model. Here, we note that the theoretical problem we consider can be seen as a discrete-time version of the classical Merton's portfolio problem \citep{merton1969lifetime}. The more flexible Heston model \citep{heston1993closed} takes the form
        $dS_t  = r S_t dt  + \sqrt{\nu_t} S_t dW_t$, 
where $\nu_t$ is the instantaneous volatility that follows its own random walk, and $dW_t$ is drawn from a Gaussian distribution. Despite the simplicity of these models, the statistical properties of these models agree well with the known statistical properties of the real financial markets \citep{dragulescu2002probability}. The readers are referred to \citep{karatzas1998methods} for a detailed discussion about the meaning and financial significance of these models.

\vspace{0mm}
\subsection{No Data Augmentation}
\vspace{0mm}
In practice, there is no way to observe more than one data point for a given stock at a given time $t$. This means that it can be very risky to directly train on the raw observed data since nothing prevents the model from overfitting to the data. Without additional assumptions, the risk is zero because there is no randomness in the training set conditioning on the time $t$. To control this risk, we thus need data augmentation. One can formalize this intuition through the following proposition, whose proof is given in Section~\ref{app: no augmentation proof}.
\begin{proposition}\label{prop: no augmentation}
{\rm (Utility of no-data-augmentation strategy.)} Let the price trajectory be generated with GBM in Eq.~\eqref{eq: geometric brownian motion} with initial price $S_0$, then the true utility for the no-data-augmentation strategy is 
\begin{equation}
    U_{\rm no-aug} =[1 - 2\Phi({-r}/{\sigma})]r - \frac{\lambda}{2} \sigma^2
\end{equation}
where $U(\pi)$ is the utility function defined in Eq.~\eqref{eq: portfolio objective}; $\Phi$ is the c.d.f. of a standard normal distribution.
\end{proposition}

This means that, the larger the volatility $\sigma$, the smaller is the utility of the no-data-augmentation strategy. This is because the model may easily overfit to the data when no data augmentation is used. In the next section, we discuss the case when a simple data augmentation is used.

\vspace{0mm}
\subsection{Additive Gaussian Noise}
\vspace{0mm}
While it is still far from clear how the stock price is correlated with the past prices, it is now well-recognized that $\text{Var}_{S_t}[S_t|S_{t-1}]\neq 0$ \citep{mandelbrot1997variation, cont2001empirical}. This motivates a simple data augmentation technique to add some randomness to the financial sequence we observe, $\{S_1,...,S_{T+1}\}$. This section analyzes a vanilla version of data augmentation of injecting simple Gaussian noise, compared to a more sophisticated data augmentation method in the next section. Here, we inject random Gaussian noises $\epsilon_t\sim \mathcal{N}(0, \rho^2)$ to $S_t$ during the training process such that $z_t = S_t + \epsilon$. Note that the noisified return needs to be carefully defined since noise might also appear in the denominator, which may cause divergence; to avoid this problem, we define the noisified return to be $\tilde{r}_t := \frac{z_{t+1} - z_t}{S_t}$, i.e., we do not add noise to the denominator. 
Theoretically, we can find the optimal strength $\rho^*$ of the gaussian data augmentation to be such that the true utility function is maximized for a fixed training set. 
The result can be shown to be
\begin{equation}\label{eq: add strength}
    (\rho^*)^2 = 
        \frac{\sigma^2}{2 r}\frac{\sum_t (r_t S_t^2)^2}{\sum_t r_t S_t^2}. 
\end{equation}
The fact the $\rho^*$ depends on the prices of the whole trajectory reflects the fact that time-independent data augmentation is not suitable for a stock price dynamics prescribed by Eq.~\eqref{eq: geometric brownian motion}, whose inherent noise $\sigma S_t \eta_t$ is time-dependent through the dependence on $S_t$. 
Finally, we can plug in the optimal $\rho^*$ to obtain the optimal achievable strategy for the additive Gaussian noise augmentation. 
As before, the above discussion can be formalized, with the true utility given in the next proposition (proof in Section~\ref{app sec: additive proof}).

\begin{proposition}{\rm (Utility of additive Gaussian noise strategy.)} Under additive Gaussian noise strategy, and let other conditions the same as in Proposition~\ref{prop: no augmentation}, the true utility is
\begin{equation}
    U_{\rm Add} = \frac{r^2}{2\lambda \sigma^2 T} \mathbb{E}_{S_t} \left[\frac{(\sum_t r_t S_t)^2}{\sum_t (r_t S_t)^2}\Theta\left(\sum_t r_t S_t^2 \right) \right],
\end{equation}
where $\Theta$ is the Heaviside step function.
\end{proposition}

\vspace{0mm}
\subsection{Multiplicative Gaussian Noise}
\vspace{0mm}
In this section, we derive a general kind of data augmentation for the price trajectories specified by the GBM and the Heston model. From the previous discussions, one might expect that a better kind of augmentation should have $\rho = \rho_0 S_t$, i.e., the injected noise should be \textit{multiplicative}; however, we do not start from imposing $\rho \to \rho S_t $; instead, we consider $\rho \to \rho_t$, i.e., a general time-dependent noise. 
In the derivation, one can find an interesting relation for the optimal augmentation strength:
\begin{equation}\label{eq: multi strength}
    (\rho_{t+1}^*)^2 +(\rho_{t}^*)^2 = 
         \frac{\sigma^2}{2r}r_t S_t^2. 
\end{equation}
The following proposition gives the true utility of using this data augmentation (derivations in Section~\ref{app sec: general mult noise proof}).
\begin{proposition}{\rm (Utility of general multiplicative Gaussian noise strategy.)} Under general multiplicative noise augmentation strategy, and let other conditions the same as in Proposition~\ref{prop: no augmentation}, then the true utility is
\begin{equation}
    U_{\rm mult} = \frac{r^2}{2\lambda \sigma^2} [1-\Phi(-r/\sigma)].
\end{equation}
\end{proposition}

Combining the above propositions, we can prove the main theorem of this work ((Proof in Section~\ref{app sec: proof of remark})), which shows that the mean-variance utility of the proposed method is strictly higher than that of no data-augmention and that of additive Gaussian noise.
\begin{theorem}\label{theo: remark}
If $\sigma \neq 0$, then $U_{\rm mult} > U_{\rm add}$ and $U_{\rm mult} > U_{\rm no-aug}$ with probability $1$. 
\end{theorem}

\textit{Heston Model and Real Price Augmentation}. We also consider the more general Heston model. The derivation proceeds similarly by replacing $\sigma^2 \to \nu_t^2$; one arrives at the relation for optimal augmentation: $(\rho_{t+1}^*)^2 +(\rho_{t}^*)^2 = \frac{1}{2r}\nu^2_tr_t S_t^2$. One quantity we do not know is the volatility $\nu_t$, which has to be estimated by averaging over the neighboring price returns. One central message from the above results is that one should add noises with variance proportional to $r_t S_t^2$ to the observed prices for augmenting the training set. 

\vspace{0mm}
\subsection{Stationary Portfolio}\label{sec: stationary portfolio}
\vspace{0mm}
In the previous sections, we have discussed the case when the portfolio is dynamic (time-dependent). One slight limitation of the previous theory is that one can only compare the in-sample counterfactual performance of a dynamic portfolio. Here, we alternatively motivate the proposed data augmentation technique when the model is a stationary portfolio. One can show that, for a stationary portfolio, the proposed data augmentation technique gives the overall optimal performance.
\begin{theorem}
    Under the multiplicative data augmentation strategy, the in-sample counterfactual utility and the out-of-sample utility is optimal among all stationary portfolios.
\end{theorem}
\begin{remark}
   See Section~\ref{app sec: stationary portfolio} for a detailed discussion and the proof. Stationary portfolios are important in financial theory and can be shown to be optimal even among all dynamic portfolios in some situations \citep{CoverInformationTheory, merton1969lifetime}. While restricting to stationary portfolios allows us to also compare on out-of-sample performance, the limitation is that a stationary portfolio is less relevant for a deep learning model than the dynamical portfolios considered in the previous sections. 
\end{remark}


\vspace{0mm}
\subsection{General Framework}\label{sec: general theory}
\vspace{0mm}
So far, we have been analyzing the data augmentation for specific examples of the utility function and the data augmentation distribution to argue that certain types of data augmentation is preferable. Now we outline how this formulation can be generalized to deal with a wider range of problems, such as different utility functions and different data augmentations. This general framework can be used to derive alternative data augmentations schemes if one wants to maximize other financial metrics other than the Sharpe ratio, such as the Sortino ratio \citep{estrada2006downside}, or to incorporate regularization effect that into account of the heavy tails of the prices distribution.

For a general utility function $U=U(x, \pi)$ for some data point $x$ that describes the current state of the market, and $\pi$ that describes our strategy in this market state, we would like to ultimately maximize
\begin{equation}
    \max_\pi V(\pi), \quad \text{for } V(\pi) =\mathbb{E}_x [U(x,\pi)]
\end{equation}
However, only observing finitely many data points, we can only optimize the empirical loss with respect to some $\theta-$parametrized augmentation distribution $P_\theta$:
\begin{equation}
    \hat{\pi} (\theta) =  \arg\max_\pi \frac{1}{N}\sum_i^N \mathbb{E}_{z_i\sim p_\theta(z|x_i)} [U(z_i,\pi_i)].
\end{equation}
The problem we would like to solve is to find the effect of using such data augmentation on the true utility $V$, and then, if possible, compare different data augmentations and identify the better one. Surprisingly, this is achievable since $V=V(\hat{\pi}(\theta))$ is now also dependent on the parameter $\theta$ of the data augmentation. Note that the true utility has to be found with respect to both the sampling over the test points and the sampling over the $N$-sized training set:
\begin{equation}\label{eq: risk}
    V(\hat{\pi}(\theta)) = \mathbb{E}_{x\sim p(x)} \mathbb{E}_{\{x_i\}^N \sim p^N(x)} [U(x,\hat{\pi}(\theta))]
\end{equation}
In principle, this allows one to identify the best data augmentation for the problem at hand:
\begin{align}
    &\theta^* = \arg\max_{\theta} V(\hat{\pi}(\theta)) \arg\max_{\theta} \mathbb{E}_{x\sim p(x)} \mathbb{E}_{\{x_i\}^N \sim p^N(x)}\nonumber\\
    &\left[U \left(x,\arg\max_\pi \frac{1}{N}\sum_i^N \mathbb{E}_{z_i\sim p_\theta(z|x_i)} [U(z_i,\pi_i)] \right)\right],
\end{align}
and the analysis we performed in the previous sections is simply a special case of obtaining solutions to this maximization problem. Moreover, one can also compare two different parametric augmentation distributions; let their parameter be denoted as $\theta_\alpha$ and $\theta_\beta$ respectively, then we can say that data augmentation $\alpha$ is better than $\beta$ if and only if
    $\max_{\theta_\alpha} V(\hat{\pi}(\theta_\alpha)) > \max_{\theta_\beta} V(\hat{\pi}(\theta_\beta)).$
This general formulation can also have applicability outside the field of finance because one can interpret the utility $U$ as a standard machine learning loss function and $\pi$ as the model output. This procedure also mimics the procedure of finding a Bayes estimator in the statistical decision theory \citep{wasserman2013all}, with $\theta$ being the estimator we want to find; we outline an alternative general formulation to find the ``minimax" augmentation in Section~\ref{app sec: analogy}.

\begin{figure*}
    \centering
    \includegraphics[width=0.37\linewidth]{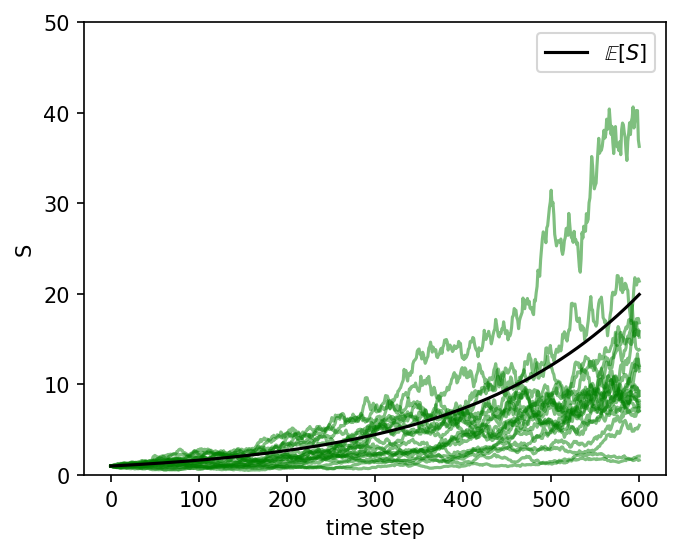}
    \includegraphics[width=0.4\linewidth]{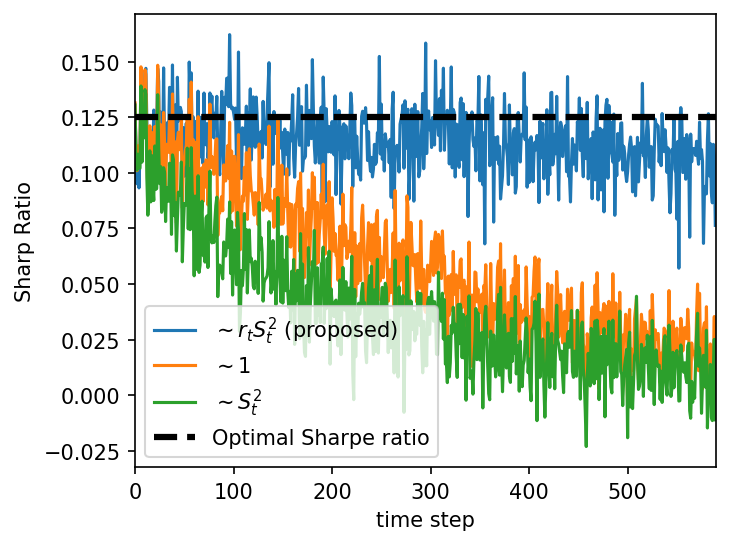}
    \vspace{-1em}
    \caption{\small Experiment on geometric brownian motion; $S_0=1$, $r=0.005$, $\sigma=0.04$. \textbf{Left}: Examples of prices trajectories in green; the black line shows the expected value of the price. \textbf{Right}: Comparison with other related data augmentation techniques. The black dashed line shows the optimal achievable Sharpe ratio. We see that the proposed method stay close to the optimality across a 600-step trading period as the theory predicts.} 
    \label{fig:geometric brownian motion}
    \vspace{-1em}
\end{figure*}

\vspace{0mm}
\section{Algorithms}
\vspace{0mm}
Our results strongly motivate for a specially designed data augmentation for financial data. For a data point consisting purely of past prices $(S_t, ..., S_{t+L}, S_{t+L +1})$ and the associated returns $(r_t, ...,r_{t + L -1}, r_{t+L})$, we use $x = (S_t, ..., S_{t+L})$ as the input for our model $f$, possibly a neural network, and use $S_{t+L +1}$ as the unseen future price for computing the training loss. Our results suggests that we should randomly noisify both the input $x$ and $S_{t+L +1}$ at every training step by 
\begin{equation}
   \begin{cases}
       S_i \to S_i +  c \sqrt{\hat{\sigma}^2_i |r_i| S_i^2} \epsilon_i \quad\quad\text{for } S_i \in x; \\
       S_{t+L+1} \to S_{t+L +1} +  c\sqrt{ \hat{\sigma}^2_i |r_{t+L}| S_{t+L}^2} \epsilon_{t+L+1}; 
   \end{cases} 
\end{equation}
where $\epsilon_i$ are i.i.d. samples from $\mathcal{N}(0,1)$, and $c$ is a hyperparameter to be tuned. While the theory suggests that $c$ should be $1/2$, it is better to make it a tunable-parameter in algorithm design for better flexibility; $\hat{\sigma}_t$ is the instantaneous volatility, which can be estimated using standard methods in finance \citep{degiannakis2015methods}. One might also assume $\hat{\sigma}$ into $c$.

\vspace{0mm}
\subsection{Using return as inputs}
\vspace{0mm}
Practically and theoretically, it is better and standard to use the returns $x=(r_t, ...,r_{t + L -1}, r_{t+L})$ as the input, and the algorithm can be applied in a simpler form:
\begin{equation}
   \begin{cases}
       r_i \to r_i +  c \sqrt{\hat{\sigma}^2_i |r_i|} \epsilon_i & \text{for } r_i \in x; \\
       r_{t+L} \to r_{t+L} +  c\sqrt{ \hat{\sigma}^2_i |r_{t+L}|} \epsilon_{t+L+1}.
   \end{cases} 
\end{equation}

\vspace{0mm}
\subsection{Equivalent Regularization on the output}
\vspace{0mm}
One additional simplification can be made by noticing the effect of injecting noise to $r_{t+L}$ on the training loss is equivalent to a regularization. We show in Section~\ref{app sec: derivation of algorithm} that, under the GBM model, the training objective can be written as
\begin{equation}\label{eq: proposed algorithm}
    \arg\max_{b_t} \left\{ \frac{1}{T} \sum_{t=1}^{T} \mathbb{E}_{z} \left[G_t(\pi) \right] - \lambda  c^2\hat{\sigma}_t^2 |r_t|  \pi_t^2 \right\},
\end{equation}
where the expectation over $x$ is now only taken with respect to the input. This means that the noise injection on the $r_{t+L}$ is equivalent to adding a $L_2$ regularization on the model output $\pi_t$. This completes the main proposed algorithm of this work. We discuss a few potential variants in Section~\ref{app sec: derivation of algorithm}. Also, it is well known that the magnitude of $|r_t|$ has strong time-correlation (i.e., a large $|r_t|$ suggests a large $|r_{t+1}|$) \citep{lux2000volatility, cont1997herd, cont2007volatility}, and this suggests that one can also use the average of the neighboring returns to smooth the $|r_t|$ factor in the last term for some time-window of width $\tau$: $|r_t| \to |\hat{r}_t| = \frac{1}{\tau}\sum_0^{\tau} |r_{t-\tau}|$. In our S\&P500 experiments, we use this smoothing technique with $\tau=20$.

\begin{table*}[t!]
\centering
{\scriptsize
    \caption{Sharpe ratio on S\&P 500 by sectors; the larger the better. Best performances in \textbf{Bold}.}\label{tab:sp500 results}
    \vspace{-1em}
    \centering
    \begin{tabular}{c|c| cccccc}
    \hline
    \hline
         Industry Sectors & \# Stock& Merton& no aug. & weight decay & additive aug. & naive mult. & proposed \\
         \hline
Communication Services& 9 &  $-0.06_{\pm 0.04} $& $-0.06_{\pm 0.04} $&  $-0.06_{\pm 0.27} $& $\mathbf{0.22_{\pm 0.18}} $ & $\mathbf{0.20_{\pm 0.21}} $ &  $\mathbf{ 0.33_{\pm 0.16}} $ \\    
Consumer Discretionary& 39 &  $-0.01_{\pm 0.03} $& $-0.07_{\pm 0.03} $&  $-0.06_{\pm 0.10} $& $0.48_{\pm 0.10} $ & $0.41_{\pm 0.09} $ &  $\mathbf{ 0.64_{\pm 0.08}} $ \\    
Consumer Staples& 27 &  $0.05_{\pm 0.03} $& $0.24_{\pm 0.03} $&  $0.23_{\pm 0.11} $& $\mathbf{0.36_{\pm 0.08}} $ & $\mathbf{0.34_{\pm 0.09}} $ &  $\mathbf{ 0.35_{\pm 0.07}} $ \\    
Energy& 17 &  $0.07_{\pm 0.03} $& $0.03_{\pm 0.03} $&  $-0.02_{\pm 0.12} $& $0.70_{\pm 0.09} $ & $0.52_{\pm 0.10} $ &  $\mathbf{ 0.91_{\pm 0.10}} $ \\    
Financials& 46 &  $-0.57_{\pm 0.04} $& $-0.61_{\pm 0.03} $&  $-0.61_{\pm 0.09} $& $-0.06_{\pm 0.10} $ & $-0.13_{\pm 0.09} $ &  $\mathbf{ 0.18_{\pm 0.08}} $ \\    
Health Care& 44 &  $0.23_{\pm 0.04} $& $0.60_{\pm 0.04} $&  $0.61_{\pm 0.11} $& $\mathbf{0.86_{\pm 0.09}} $ & $\mathbf{0.81_{\pm 0.09}} $ &  $\mathbf{ 0.83_{\pm 0.07}} $ \\    
Industrials& 44 &  $-0.09_{\pm 0.03} $& $-0.11_{\pm 0.03} $&  $-0.11_{\pm 0.08} $& $0.36_{\pm 0.08} $ & $0.28_{\pm 0.08} $ &  $\mathbf{ 0.48_{\pm 0.08}} $ \\    
Information Technology& 41 &  $0.41_{\pm 0.04} $& $0.41_{\pm 0.04} $&  $0.41_{\pm 0.11} $& $0.67_{\pm 0.10} $ & $\mathbf{0.74_{\pm 0.11}} $ &  $\mathbf{ 0.79_{\pm 0.09}} $ \\    
Materials& 19 &  $0.07_{\pm 0.03} $& $0.06_{\pm 0.03} $&  $0.03_{\pm 0.14} $& $\mathbf{0.47_{\pm 0.13}} $ & $\mathbf{0.43_{\pm 0.13}} $ &  $\mathbf{ 0.53_{\pm 0.10}} $ \\    
Real Estate& 22 &  $-0.14_{\pm 0.04} $& $-0.39_{\pm 0.03} $&  $-0.40_{\pm 0.12} $& $0.05_{\pm 0.10} $ & $0.05_{\pm 0.09} $ &  $\mathbf{ 0.19_{\pm 0.07}} $ \\    
Utilities& 24 &  $-0.29_{\pm 0.02} $& $-0.29_{\pm 0.02} $&  $-0.28_{\pm 0.07} $& $-0.01_{\pm 0.06} $ & $-0.00_{\pm 0.06} $ &  $\mathbf{ 0.15_{\pm 0.04}} $ \\    
\hline
        S\&P500 Avg.& 365 & $-0.02_{\pm 0.04} $ & $-0.00_{\pm 0.04} $&  $-0.01_{\pm 0.04} $& $0.39_{\pm 0.03} $ & $0.35_{\pm 0.03} $ &  $\mathbf{0.51_{\pm 0.03}}$\\
         \hline
    \end{tabular}}
    \vspace{-1em}
\end{table*}

 \vspace{0mm}
\section{Experiments}
\vspace{0mm}
We validate our theoretical claim that using multiplicative noise with strength $\sqrt{r}$ is better than not using any data augmentation or using a data augmentation that is not suitable for the nature of portfolio construction (such as an additive Gaussian noise). We emphasize that the purpose of this section is for demonstrating the relevance of our theory to real financial problems, not for establishing the proposed method as a strong competitive method in the industry. We start with a toy dataset that follows the theoretical assumptions and then move on to real data with S\&P500 prices. The detailed experimental settings are given in Section~\ref{app sec: exp detail}. Unless otherwise specified, we use a feedforward neural network with the number of neurons $10 \to 64 \to 64 \to 1$ with ReLU activations. Training proceeds with the Adam optimizer with a minibatch size of $64$ for 100 epochs with the default parameter settings.\footnote{In our initial experiments, we also experimented with different architectures (different depth or width of the FNN, RNN, LSTM), and our conclusion that the proposed augmentation outperforms the specified baselines remain unchanged.}

We use the Sharpe ratio as the performance metric (the larger the better). Sharpe ratio is defined as
    $SR_t = \frac{\E [\Delta W_t]}{\sqrt{\text{Var}[\Delta W_t]}}$, 
which is a measure of the profitability per risk. We choose this metric because, in the framework of portfolio theory, it is the only theoretically motivated metric of success \citep{sharpe1966mutual}. In particular, our theory is based on the maximization of the mean-variance utility in Eq.~\eqref{eq: portfolio objective} and it is well-known that the maximization of the mean-variance utility is equivalent to the maximization of the Sharpe ratio. In fact, it is a classical result in classical financial research that all optimal strategies must have the same Sharpe ratio \citep{sharpe1964capital} (also called the efficient capital frontier). For the synthetic tasks, we can generate arbitrarily many test points to compare the Sharpe ratios unambiguously. We then move to experiments on real stock price series; the limitation is that the Sharpe ratio needs to be estimated and involves one additional source of uncertainty.\footnote{We caution the readers to not to confuse the problem of portfolio construction with the problem of financial price prediction. Portfolio construction is the primary focus of our work and \textit{is fundamentally different from the problem of financial price prediction}. Our method is not relevant and cannot be used directly for predicting future prices. As in real life, one does not need to be able to predict prices to decide which stock to purchase.}

\vspace{0mm}\vspace{0mm}
\subsection{Geometric Brownian Motion}\label{sec: gbm experiments}
\vspace{0mm}
We first start from experimenting with stock prices generated with a GBM, as specified in Eq.~\eqref{eq: geometric brownian motion}, and we generate a fixed price trajectory with length $T=400$ for training; each training point consists of a sequence of past prices $(S_t, ..., S_{t+9}, S_{t+10})$ where the first ten prices are used as the input to the model, and $S_{t+10}$ is used for computing the loss.

\textit{Results and discussion}. See Figure~\ref{fig:geometric brownian motion}. The proposed method is plotted in {\color{blue} blue}. The right figure compares the proposed method with the other two baseline data augmentations we studied in this work. As the theory shows, the proposed method is optimal for this problem, achieving the optimal Sharpe ratio across a 600-step trading period. This directly confirms our theory.

\vspace{0mm}\vspace{0mm}
\subsection{S\&P 500 Prices}
\vspace{0mm}
This section demonstrates the relevance of the proposed algorithm to real market data. In particular, We use the data from S\&P500 from 2016 to 2020, with $1000$ days in total. We test on the $365$ stocks that existed on S\&P500 from $2000$ to $2020$. We use the first $800$ days as the training set and the last $200$ days for testing. The model and training setting is similar to the previous experiment. 
We treat each stock as a single dataset and compare on all of the $365$ stocks (namely, the evaluation is performed independently on $365$ different datasets). Because the full result is too long, We report the average Sharpe ratio per industrial sector (categorized according to GISC) and the average Sharpe ratio of all 365 datasets. See Section~\ref{acc sec: exp detail} and \ref{acc sec: sp500 exp} for more detail.

\textit{Results and discussion}. See Table~\ref{tab:sp500 results}. We see that, without data augmentation, the model works poorly due to its incapability of assessing the underlying risk. We also notice that weight decay does not improve the performance (if it is not deteriorating the performance). We hypothesize that this is because weight decay does not correctly capture the inductive bias that is required to deal with a financial series prediction task. Using any kind of data augmentation seems to improve upon not using data augmentation. Among these, the proposed method works the best, possibly due to its better capability of risk control. In this experiment, we did not allow for short selling; when short selling is allowed, the proposed method also works the best; see Section~\ref{acc sec: sp500 exp}. In Section~\ref{app sec: case study}, we also perform a case study to demonstrate the capability of the learned portfolio to avoid a market crash in 2020. We also compare with the Merton's portfolio \citep{merton1969lifetime}, which is the classical optimal stationary portfolio constructed from the training data; this method does not perform well either. This is because the market during the time $2019-2020$ is volatile and quite different from the previous years, and a stationary portfolio cannot capture the nuances in the change of the market condition. This shows that it is also important to leverage the flexibility and generalization property of the modern neural networks, along side the financial prior knowledge. 

\begin{figure}
    \centering
    \includegraphics[width=0.3\linewidth]{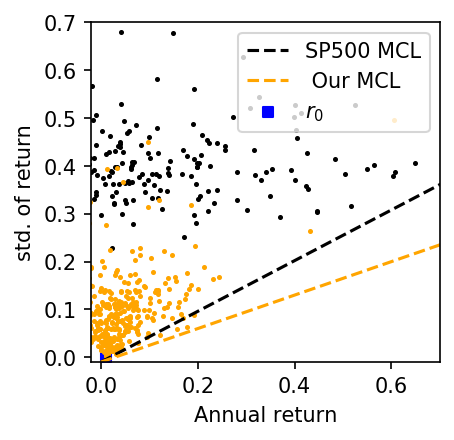}
    \vspace{-1em}
    \caption{\small Available portfolios and the market capital line (MCL). The \textbf{black} dots are the return-risk combinations of the original stocks; the {\color{orange}\textbf{orange}} dots are the learned portfolios. The MCL of the proposed method is lower than that of the original stocks, suggesting improved return and lower risk.}
    \label{fig:market capital line}
    \vspace{-1em}
\end{figure}

\subsection{Comparison with Data Generation Method}

One common alternative to direct data augmentation in the field is to generate additional realistic synthetic data using a GAN. While it is not the purpose of this work to propose an industrial level method, nor do we claim that the proposed method outperforms previous methods, we provide one experimental comparison in Section~\ref{app sec: gan comparison} for the task of portfolio construction. We compare our theoretically motivated technique with QuantGAN \citep{wiese2020quant}, a major and recent technique in the field of financial data augmentation/generation. The experiment setting is the same as the S\&P500 experiment. The result shows that directly applying QuantGAN to the portfolio construction problem in our setting does not significantly improve over the baseline without any augmentation and achieves a much lower Sharpe ratio than our suggested method. This underperformance is possibly because QuantGAN is not designed for Sharpe ratio maximization.

\vspace{0mm}
\subsection{Market Capital Lines}
\vspace{0mm}
In this section, we link the result we obtained in the previous section with the concept of market capital line (MCL) in the capital asset pricing model \citep{sharpe1964capital}, a foundational theory in classical finance. The MCL of a set of portfolios denotes the line of the best return-risk combinations when these portfolios are combined with a risk-free asset such as the government bond; an MCL with smaller slope means better return and lower risk and is considered to be better than an MCL that is to the upper left in the return-risk plane. See Figure~\ref{fig:market capital line}. The risk-free rate $r_0$ is set to be $0.01$, roughly equal to the average 1-year treasury yield from 2018 to 2020. We see that the learned portfolios achieves a better MCL than the original stocks. The slope of the SP500 MCL is roughly $0.53$, while that of the proposed method is $0.35$, i.e., much better return-risk combinations can be achieved using the proposed method. For example, if we specify the acceptable amount of risk to be $0.1$, then the proposed method can result in roughly $10\%$ more gain in annual return than investing in the best stock in the market.
This example also shows that how tools in classical finance theory can be used to visualize and better understand the machine learning methods that are applied to finance, a crucial point that many previous works lack.

\subsection{Case Study}

For completeness, we also present the performance of the proposed method during the Market crush in Feb. 2020 for the interested readers. See Section~\ref{app sec: case study}.

\section{Outlook}\label{sec: outlook}

In this work, we have presented a theoretical framework relevant to finance and machine learning to understand and analyze methods related to deep-learning-based finance. The result is a machine learning algorithm incorporating prior knowledge about the underlying financial processes. The good performance of the proposed method agrees with the standard expectation in machine learning that performance can be improved if the right inductive biases are incorporated. We have thus succeeded in showing that building machine learning algorithms that is rooted firmly in financial theories can have a considerable and yet-to-be achieved benefit. We hope that our work can help motivating for more works that approaches the theoretical aspects of machine learning algorithms that are used for finance.

The limitation of the present work is obvious; we only considered the kinds of data augmentation that takes the form of noise injection. Other kinds of data augmentation may also be useful to the finance; for example, \citep{fons2020evaluating} empirically finds that magnify \citep{um2017data}, time warp \citep{kamycki2020data}, and SPAWNER \citep{le2016data} are helpful for financial series prediction, and it is interesting to apply our theoretical framework to analyze these methdos as well; a correct theoretical analysis of these methods is likely to advance both the deep-learning based techniques for finance and our fundamental understanding of the underlying financial and economic mechanisms. Meanwhile, our understanding of the underlying financial dynamics is also rapidly advancing; we foresee better methods to be designed, and it is likely that the proposed method will be replaced by better algorithms soon. There is potentially positive social effects of this work because it is widely believed that designing better financial prediction methods can make the economy more efficient by eliminating arbitrage \citep{fama1970efficient}; the cautionary note is that this work is only for the purpose of academic research, and should not be taken as an advice for monetary investment, and the readers should evaluate their own risk when applying the proposed method.


\appendix

\onecolumn
\section{Experiments}\label{app sec: exp detail}
This section describes the additional experiments and the experimental details in the main text. The experiments are all done on a single TITAN RTX GPU. The S\&P500 data is obtained from Alphavantage.\footnote{{https://www.alphavantage.co/documentation/}} The code will be released on github.

\subsection{Dataset Construction}\label{acc sec: exp detail}
For all the tasks, we observe a single trajectory of  a single stock prices $S_1, ..., S_T$. For the toy tasks, $T=400$; for the S\&P500 task, $T=800$. We then transform this into $T-L$ input-target pairs $\{(x_i,y_i)\}_{i=1}^{T-L}$, where 
\begin{equation}
    \begin{cases}
        x_i = (S_i,...,S_{L-1});\\
        y_i = S_{L}.
    \end{cases}
\end{equation}
$x_i$ is used as the input to the model for training; $y_i$ is used as the unseen future price for calculating the loss function. For the toy tasks, $L=10$; for the S\&P500 task, $L=15$. In simple words, we use the most recent $L$ prices for constructing the next-step portfolio.

\subsection{Sharpe Ratio for S\&P500}
The empirical Sharpe Ratios are calculated in the standard way (for example, it is the same as in \citep{ito2020tradercompany, imajo2020deep}). Given a trajectory of wealth $W_1,...,W_T$ of a strategy $\pi$, the empirical Sharpe ratio is estimated as
\begin{equation}
    R_i = \frac{W_{i+1}}{W_i} - 1;
\end{equation}
\begin{equation}
    \hat{M} = \frac{1}{T}\sum_{i=1}^{T-1} R_i;
\end{equation}
\begin{equation}
    \widehat{SR} =  \frac{ \hat{M}}{\sqrt{\frac{1}{T} \sum_{i=1}^T R_i^2 - \hat{M}^2}} = \frac{\text{average wealth return}}{\text{std. of wealth return}},
\end{equation}
and $\widehat{SR}$ is the reported Sharpe Ratio for S\&P500 experiments.

\subsection{Variance of Sharpe Ratio}
We do not report an uncertainty for the single stock Sharpe Ratios, but one can easily estimate the uncertainties. The Sharpe Ratio is estimated across a period of $T$ time steps. For the $S\&P500$ stocks, $T=200$, and by the law of large numbers, the estimated mean $\hat{M}$ has variance roughly $\sigma^2/T$, where $\sigma$ is the true volatility, and so is the estimated standard deviation. Therefore, the estimated Sharpe Ratio can be written as
\begin{align}
    \widehat{SR} &= \frac{ \hat{M}}{\sqrt{\frac{1}{T} \sum_{i=1}^T R_i^2 - \hat{M}^2}}\\
    &= \frac{ M + \frac{\sigma}{\sqrt{T}}\epsilon}{\sigma + \frac{c}{\sqrt{T}}\eta}\\
    & \approx \frac{ M + \frac{\sigma}{\sqrt{T}}\epsilon}{\sigma } = \frac{M}{\sigma} + \frac{1}{\sqrt{T}} \epsilon
\end{align}
where $\epsilon$ and $\eta$ are zero-mean random variables with unit variance. This shows that the uncertainty in the estimated $\widehat{SR}$ is approximately $1/\sqrt{T} \approx 0.07$ for each of the single stocks, which is often much smaller than the difference between different methods.

\subsection{S\&P500 Experiment}\label{acc sec: sp500 exp}
This section gives more results and discussion of the $S\&P500$ experiment.
\subsubsection{Underperformance of Weight Decay}
This section gives the detail of the comparison made in Figure~\ref{fig: motivation}. The experimental setting is the same as the $S\&P500$ experiments. For illustration and motivation, we only show the result on MSFT (Microsoft). Choosing most of the other stocks would give a qualitatively similar plot.  

See Figure~\ref{fig: motivation}, where we show the performance of directly training a neural network to maximize wealth return on MSFT during 2018-2020. Using popular, generic deep learning techniques such as weight decay or dropout does not improve the baseline. In contrast, our theoretically motivated method does. Combining the proposed method with weight decay has the potential to improve the performance a little further, but the improvement is much lesser than the improvement of using the proposed method over the baseline. This implies that generic machine learning is unlikely to capture the inductive bias required to process a financial task.

In the plot, we did not interpolate the dropout method between a large $p$ and a small $p$. The result is similar to the case of weight decay in our experiments.

\begin{figure}
    \centering
    \includegraphics[width=0.4\linewidth]{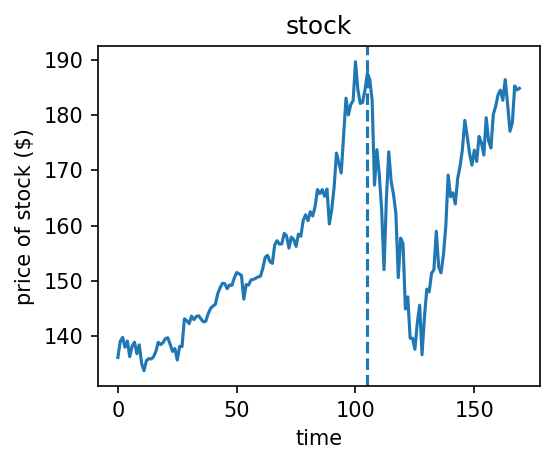}
    \includegraphics[width=0.5\linewidth]{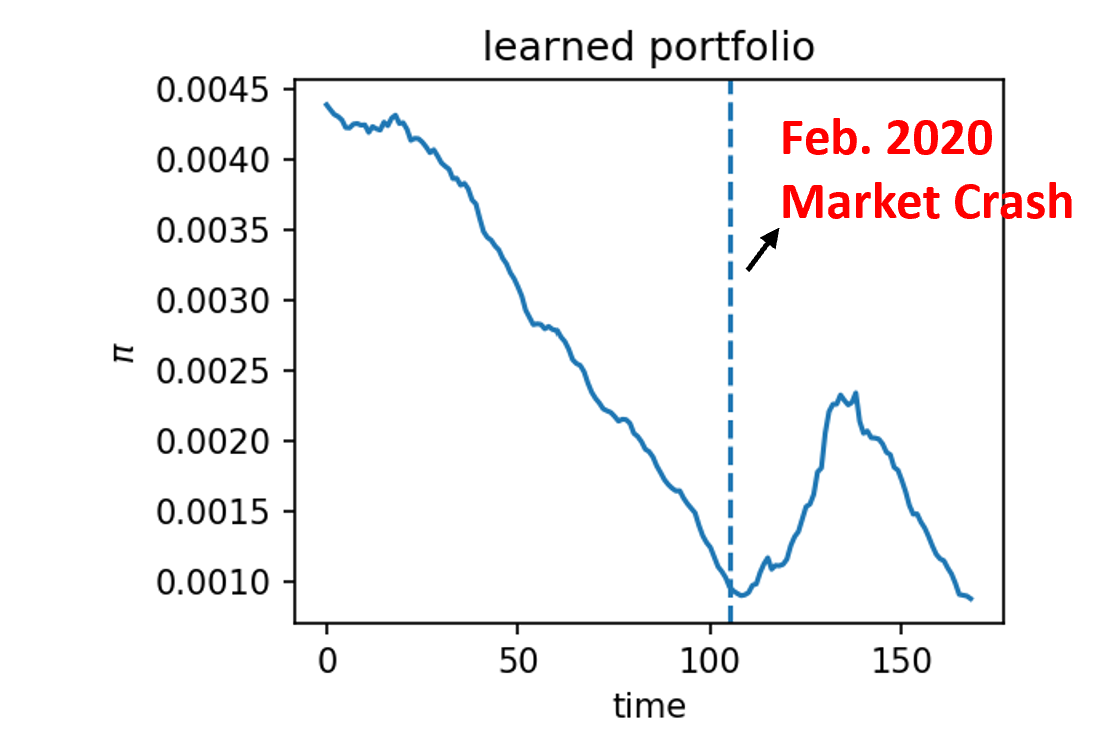}
    \caption{Case study of the performance of the model on MSFT from 2019 August to 2020 May. We see that the model learns to invest less and less as the price of the stock rises to an unreasonable level, thus avoiding the high risk of the market crash in February 2020.}
    \label{fig:case study}
\end{figure}

\subsection{Comparison with GAN}\label{app sec: gan comparison}
\begin{wrapfigure}{r}{0.4\textwidth}
\vspace{-1em}
    \centering
    {\footnotesize
    \begin{tabular}{c|ccccccc}
    Sector & Q-GAN Aug.& Ours\\
    \hline
    Comm. Services &  $0.07_{\pm 0.02} $& $0.33_{\pm 0.16}  $\\    
    Consumer Disc. &  $0.00_{\pm 0.03} $& $0.64_{\pm 0.08}  $\\    
    Consumer Staples &  $0.10_{\pm 0.02} $& $0.35_{\pm 0.07}  $\\    
    Energy &  $0.08_{\pm 0.05} $& $0.91_{\pm 0.10}  $\\    
    Financials &  $-0.22_{\pm 0.03} $& $0.18_{\pm 0.08}  $\\    
    Health Care &  $0.35_{\pm 0.03} $& $0.83_{\pm 0.07}  $\\    
    Industrials &  $-0.12_{\pm 0.03} $& $0.48_{\pm 0.08}  $\\    
    Information Tech. &  $0.28_{\pm 0.04} $& $0.79_{\pm 0.09}  $\\    
    Materials &  $-0.03_{\pm 0.03} $& $0.53_{\pm 0.10}  $\\    
    Real Estate &  $-0.35_{\pm 0.03} $& $0.19_{\pm 0.07}  $\\    
    Utilities &  $-0.20_{\pm 0.02} $& $0.15_{\pm 0.04}  $\\   
    \hline
    S\&P500 Avg. & $0.05_{\pm 0.03} $& $\mathbf{0.51}_{\pm 0.03}  $ 
    
    \end{tabular}
    }
    \caption{\footnotesize Performance (Sharpe Ratio) of the training set augmented with QuantGAN.}
\end{wrapfigure}
We stress that it is not the goal of this paper to compare methods but to understand why certain methods work or do not work from the perspective of the classical finance theory. With this caveat clearly stated, we compare our theoretically motivated technique with QuantGAN, a major and recent technique in the field of financial data augmentation/generation. We first train a QuantGAN on each stock trajectory and use the trained QuantGAN to generate 10000 additional data points to augment the original training set (containing roughly 1000 data points for each stock) and train the same feedforward net as described in the main text. This feedforward net is then used for evaluation. QuantGAN is implemented as close to the original paper as possible, using a temporal convolutional network trained with RMSProp for 50 epochs. Other experimental settings are the same as that of S\&P500 experiment in the manuscript. See the right Table. Also, compare with Table 1 in the manuscript. We see that directly applying QuantGAN to the portfolio construction problem in our setting does not significantly improve over the baseline without any augmentation and achieves a much lower Sharpe ratio than our suggested method. This underperformance is possibly because the QuantGAN is not designed for Sharpe ratio maximization.

\subsubsection{Case Study}\label{app sec: case study}
In this section, we qualitatively study the behavior of the learned portfolio of the proposed method. The model is trained as in the other S\&P500 experiments. See Figure~\ref{fig:case study}. We see that the model learns to invest less and less as the stock price rises to an excessive level, thus avoiding the high risk of the market crash in February 2020. This avoidance demonstrates the effectiveness of the proposed method qualitatively.

\subsubsection{List of Symbols for S\&P500}\label{acc sec: sp500 list}
The following are the symbols we used for the S\&P500 experiments, separated by quotation marks.
\\

['A' 'AAPL' 'ABC' 'ABT' 'ADBE' 'ADI' 'ADM' 'ADP' 'ADSK' 'AEE' 'AEP' 'AES'
 'AFL' 'AIG' 'AIV' 'AJG' 'AKAM' 'ALB' 'ALL' 'ALXN' 'AMAT' 'AMD' 'AME'
 'AMG' 'AMGN' 'AMT' 'AMZN' 'ANSS' 'AON' 'AOS' 'APA' 'APD' 'APH' 'ARE'
 'ATVI' 'AVB' 'AVY' 'AXP' 'AZO' 'BA' 'BAC' 'BAX' 'BBT' 'BBY' 'BDX' 'BEN'
 'BIIB' 'BK' 'BKNG' 'BLK' 'BLL' 'BMY' 'BRK.B' 'BSX' 'BWA' 'BXP' 'C' 'CAG'
 'CAH' 'CAT' 'CB' 'CCI' 'CCL' 'CDNS' 'CERN' 'CHD' 'CHRW' 'CI' 'CINF' 'CL'
 'CLX' 'CMA' 'CMCSA' 'CMI' 'CMS' 'CNP' 'COF' 'COG' 'COO' 'COP' 'COST'
 'CPB' 'CSCO' 'CSX' 'CTAS' 'CTL' 'CTSH' 'CTXS' 'CVS' 'CVX' 'D' 'DE' 'DGX'
 'DHI' 'DHR' 'DIS' 'DLTR' 'DOV' 'DRE' 'DRI' 'DTE' 'DUK' 'DVA' 'DVN' 'EA'
 'EBAY' 'ECL' 'ED' 'EFX' 'EIX' 'EL' 'EMN' 'EMR' 'EOG' 'EQR' 'EQT' 'ES'
 'ESS' 'ETFC' 'ETN' 'ETR' 'EW' 'EXC' 'EXPD' 'F' 'FAST' 'FCX' 'FDX' 'FE'
 'FFIV' 'FISV' 'FITB' 'FL' 'FLIR' 'FLS' 'FMC' 'FRT' 'GD' 'GE' 'GILD' 'GIS'
 'GLW' 'GPC' 'GPS' 'GS' 'GT' 'GWW' 'HAL' 'HAS' 'HBAN' 'HCP' 'HD' 'HES'
 'HIG' 'HOG' 'HOLX' 'HON' 'HP' 'HPQ' 'HRB' 'HRL' 'HRS' 'HSIC' 'HST' 'HSY'
 'HUM' 'IBM' 'IDXX' 'IFF' 'INCY' 'INTC' 'INTU' 'IP' 'IPG' 'IRM' 'IT' 'ITW'
 'IVZ' 'JBHT' 'JCI' 'JEC' 'JNJ' 'JNPR' 'JPM' 'JWN' 'K' 'KEY' 'KLAC' 'KMB'
 'KMX' 'KO' 'KR' 'KSS' 'KSU' 'L' 'LB' 'LEG' 'LEN' 'LH' 'LLY' 'LMT' 'LNC'
 'LNT' 'LOW' 'LRCX' 'LUV' 'M' 'MAA' 'MAC' 'MAR' 'MAS' 'MAT' 'MCD' 'MCHP'
 'MCK' 'MCO' 'MDT' 'MET' 'MGM' 'MHK' 'MKC' 'MLM' 'MMC' 'MMM' 'MNST' 'MO'
 'MOS' 'MRK' 'MRO' 'MS' 'MSFT' 'MSI' 'MTB' 'MTD' 'MU' 'MYL' 'NBL' 'NEE'
 'NEM' 'NI' 'NKE' 'NKTR' 'NOC' 'NOV' 'NSC' 'NTAP' 'NTRS' 'NUE' 'NVDA'
 'NWL' 'O' 'OKE' 'OMC' 'ORCL' 'ORLY' 'OXY' 'PAYX' 'PBCT' 'PCAR' 'PCG'
 'PEG' 'PEP' 'PFE' 'PG' 'PGR' 'PH' 'PHM' 'PKG' 'PKI' 'PLD' 'PNC' 'PNR'
 'PNW' 'PPG' 'PPL' 'PRGO' 'PSA' 'PVH' 'PWR' 'PXD' 'QCOM' 'RCL' 'RE' 'REG'
 'REGN' 'RF' 'RJF' 'RL' 'RMD' 'ROK' 'ROP' 'ROST' 'RRC' 'RSG' 'SBAC' 'SBUX'
 'SCHW' 'SEE' 'SHW' 'SIVB' 'SJM' 'SLB' 'SLG' 'SNA' 'SNPS' 'SO' 'SPG'
 'SRCL' 'SRE' 'STT' 'STZ' 'SWK' 'SWKS' 'SYK' 'SYMC' 'SYY' 'T' 'TAP' 'TGT'
 'TIF' 'TJX' 'TMK' 'TMO' 'TROW' 'TRV' 'TSCO' 'TSN' 'TTWO' 'TXN' 'TXT'
 'UDR' 'UHS' 'UNH' 'UNM' 'UNP' 'UPS' 'URI' 'USB' 'UTX' 'VAR' 'VFC' 'VLO'
 'VMC' 'VNO' 'VRSN' 'VRTX' 'VTR' 'VZ' 'WAT' 'WBA' 'WDC' 'WEC' 'WFC' 'WHR'
 'WM' 'WMB' 'WMT' 'WY' 'XEL' 'XLNX' 'XOM' 'XRAY' 'XRX' 'YUM' 'ZION']

\clearpage

\section{Additional Discussion of the Proposed Algorithm}\label{app sec: derivation of algorithm}

\subsection{Derivation}
We first derive Equation~\ref{eq: proposed algorithm}. The original training loss is
\begin{align}
 \frac{1}{T} \sum_{t=1}^{T} \mathbb{E}_{t} \left[G_t(\pi) \right] - \lambda \text{Var}_{t}[ G_t(\pi)].
\end{align}
The last term can be written as
\begin{align}
    \lambda \text{Var}_{t}[ G_t(\pi)] &= \mathbb{E}_{z_1,...,z_t}[z_t^2 \pi_t^2] - \mathbb{E}_{z_1,...,z_t}[z_t \pi_t]^2 \\
    &= \mathbb{E}_{z_t}[z_t^2]\mathbb{E}_{z_1,...,z_{t-1}}[ \pi_t^2] - \mathbb{E}_{z_t}[z_t]^2\mathbb{E}_{z_1,...,z_{t-1}}[ \pi_t]^2 \\
    &= \lambda r_t^2  \mathbb{E}_{z_1,...,z_{t-1}}[\pi_t^2] + \lambda  c^2\hat{\sigma}_t^2 |r_t| \mathbb{E}_{z_1,...,z_{t-1}} [\pi_t^2]  - \lambda r_t^2  \mathbb{E}_{z_1,...,z_{t-1}}[\pi_t]^2\\
    &= \lambda r_t^2 \text{Var}_{z_1,...,z_{t-1}}[\pi_t] + \lambda  c^2\hat{\sigma}_t^2 |r_t| \mathbb{E}_{z_1,...,z_{t-1}} [\pi_t^2]
\end{align}

Plug in, this leads to the following maximization problem, which is the desired equation.
\begin{equation}\label{eq: full training objective}
    \arg\max_{b_t} \left\{ \frac{1}{T} \sum_{t=1}^{T} \underbrace{\mathbb{E}_{x} \left[G_t(\pi) \right]}_{A:\text{ wealth gain}} - \underbrace{\lambda r_t^2 \text{Var}_{z_1,...,z_{t-1}}[\pi_t] }_{B:\text{ Risk due to uncertainty in past price}} - \underbrace{\lambda  c^2\hat{\sigma}_t^2 |r_t| \mathbb{E}_{z_1,...,z_{t-1}} [\pi_t^2]}_{C:\text{ Risk due to Future Price}} \right\},
\end{equation}
where we have given each term a name for reference; the expectation is taken with respect to the augmented data points $z_i$:
\begin{equation}
       r_i \to  z_i = r_i +  c \sqrt{\hat{\sigma}^2_i |r_i|} \epsilon_i \quad \text{for } r_i \in x.
\end{equation}

Under the GBM model (or when the optimal portfolio only weakly depends on $z_1,...,z_{t-1}$), the optimal $\pi_t$ does not depend on $z_1,...,z_{t-1}$, and so the objective can be further simplified to be 
\begin{equation}
    \arg\max_{b_t} \left\{ \frac{1}{T} \sum_{t=1}^{T} \mathbb{E}_{z} \left[G_t(\pi) \right] - \lambda  c^2\hat{\sigma}_t^2 |r_t|  \pi_t^2 \right\};
\end{equation}
the first term is the data augmentation for wealth gain, and the second term is a regularization for risk control. Most of the experiments in this paper use this equation for training. When it does not work well, the readers are encouraged to try the full training objective in Equation~\ref{eq: full training objective}.

\subsection{Extension to Multi-Asset Setting}
It is possible and interesting to derive the data augmentation for a multi-asset setting. However, this is hindered by the lack of a standard model to describe the co-motion of multiple stocks. For example, it is unsure what the geometric Brownian motion should be for a multi-stock setting. In this case, we tentatively suggest the following form of the formula for the injected noise, whose effectiveness and theoretical justification are left for future work. Let $\mathbf{S}_t=(S_{1,t}, ..., S_{N,t})$ be the prices of the stocks viewed as an $N$-dimensional vector. The return is assumed to have covariance $\Sigma$, then, by analogy with the discovery of this work:
\begin{equation}
    {S}_{i,t} \to  {S}_{i,t} + c \sqrt{\sum_j \Sigma_{ij}|r_j| S_j^2} \epsilon_t 
\end{equation}
for some white gaussian noise $\epsilon$ and some tunable parameter $c$. The matrix $\Sigma$ has to be estimated by the data using standard methods of estimating multi-stock volatility.

\subsection{Non-Gaussian Noise}
While the proposed algorithm proposes to use Gaussian noise for injection; the theory developed in this work only requires the noise to have a finite second moment and, therefore, any other distribution works as well for the particular kind of utility function we specified in Eq.~\eqref{eq: test objective}. This is because this utility function only contains the second moments of the wealth return and is indifferent to higher moments. Therefore, there is one caveat to the present theory: when the utility function involves higher moments of the wealth return, the utility of a certain type of noise injection is not indifferent to the choice of the injection distribution. The practitioners are recommended to analyze the specific utility function they use in our framework and decide on the strength and distribution of the injected noise.

\clearpage

\section{Additional Theory and Proofs}\label{app sec: proofs}

This section contains all the additional theoretical discussions and the proofs.

\subsection{Background: Classical Solution to the Portfolio Construction Problem}\label{app sec: classical portfolio solution}

Consider a market with $N$ stocks, we denote the price of all these stocks as ${S}_{t} \in \mathbb{R}^N$, and one can likewise define the stock price return as 
\begin{equation}
    r_t = \frac{S_{t+1} - S_{t}}{S_{t}}.
\end{equation}
which is a random variable with the covariance matrix $C_t :=  \text{Cov}[r_t]$ and the expected return $g_t := \mathbb{E}[r_t]$. Our wealth at time step $t$ is defined as 
    $W_t = M_t + \sum_i^N n_{t,i} S_{t, i} := n^{\rm T}_{t} S_t$ 
where $M_t$ is the amount of cash we hold, and $n_i$ the shares of stock we hold for the $i$-th stock; we also defined vectors $n_t:= (n_{t,i}, .., n_{t, N}, 1)$ and $S_t := (S_{t,1}, ..., S_{t, N}, M_t)$ where the cash is included in the definition of $n_t$. As in the standard finance literature, we assume that the shares are infinitely divisible; usually, a positive $n_i$ denotes holding (long) and a negative $n$ denotes borrowing (short). The wealth we hold initially is $W_0 > 0$, and we would like to invest our money on $N$ stocks; we denote the relative value of each stock we hold also as a vector $\pi_{t, i} = \frac{n_{t, i} S_{t,i}}{W_t}\in \mathbb{R}^{N+1}$; $\pi$ is called a \textit{portfolio}; the central challenge in portfolio theory is to find the best $\pi$. At time $t$, our relative wealth is $W_t$; after one time step, our wealth changes due to a change in the price of the stocks:
    $\Delta W_{t} := W_{t+1} - W_{t} = W_t \pi_t \cdot r_t$ .
The standard goal is to maximize the wealth return $G_t := \pi_t \cdot r_t$ at every time step \textit{while minimizing risk}\footnote{It is important to not to confuse the \textit{price return} $r_t$ with the wealth return $G_t$.}. The risk is defined as the variance of the wealth change\footnote{In principle, any concave function in $G_t$ can be a risk regularizer from classical economic theory \citep{von1947theory}, and our framework can be easily extended to such cases; one common alternative would be $R(G) = \log(G)$.}:
\begin{equation}
    R_t := R(\pi_t) :=  \text{Var}_{r_t}[G_t] = \pi_t^{\rm T} \left(\E[r_t r_t^{\rm T}] -g_t g_t^{\rm T} \right)\pi_t    = \pi_t^{\rm T} C_t \pi_t.
\end{equation}
 The standard way to control risk is to introduce a ``risk regularizer" that punishes the portfolios with a large risk \citep{markowitz1959portfolio, rubinstein2002markowitz}. Introducing a parameter $\lambda$ for the strength of regularization (the factor of $1/2$ appears for convention), we can now write down our objective:
\begin{equation}\label{eq: portfolio objective app}
    \pi_t^* = \arg\max_\pi U(\pi) :=  \arg\max_\pi \left[\pi^{\rm T} G_t - \frac{\lambda}{2} R(\pi)  \right].
\end{equation}
Here, $U$ stands for the utility function; $\lambda$ can be set to be the desired level of risk-aversion. When $g_t$ and $C_t$ is known, this problem can be explicitly solved (see Section~\ref{app sec: classical portfolio solution}). However, one main problem in finance is that its data is very limited and we only observe one particular \textit{realized} data trajectory, and, therefore, $g_t$ and $C_t$ cannot be accurately estimated. 

Eq.~\eqref{eq: portfolio objective app} can be solved directly by taking derivative and set to $0$; we can obtain 
\begin{equation}
    \begin{cases}
        \pi_t^* = \frac{1}{\lambda} C_t^{-1} g_t;\\
        R_t(\pi_t^*) = \frac{1}{\lambda^2} g_t^{\rm T} C_t^{-1} g_t.
    \end{cases}
\end{equation}
This formula is the standard formula to use when both $g_t$ and $C_t$ are known or can be accurately estimated \citep{bouchaud2009financial}. Meanwhile, when one finds difficulty estimating $g_t$ or, more importantly, $C_t$, then the above formula can go arbitrarily wrong. Let $\hat{C}$ denote our estimated covariance and $\hat{g}$ the estimated mean\footnote{For example, using some Bayesian machine learning model.}, then the in-sample risk and the true is respectively given by
\begin{equation}
\begin{cases}
        \hat{R}_t(\pi_t^*) = \frac{1}{\lambda^2} \hat{g}_t^{\rm T} \hat{C}_t^{-1} \hat{g}_t;\\
        {R}_t(\pi_t^*) = \frac{1}{\lambda^2} \hat{g}_t^{\rm T} \hat{C}_t^{-1} C_t \hat{C}_t^{-1}\hat{g}_t.
\end{cases}
\end{equation}
The readers are encouraged to examine the differences between the two equations carefully. 

\subsection{Analogy to Statistical Decision Theory and the Minimax Formulation}\label{app sec: analogy}
In Section~\ref{sec: general theory}, we mentioned that the procedure we used is analogous to the process of finding a Bayesian estimator in a statistical decision theory. Here, we explain this analogy a little more (but keep in mind that this is only an analogy, not a rigorous equivalence relation). Equation~\ref{eq: risk} of empirical utility can be seen as an equivalent of the statistical risk function $R$; finding the optimal data augmentation strength is similar to finding the best Bayesian estimator. To make an exact agreement with the statistical decision theory, we also need to define a prior over the risk in Equation~\ref{eq: risk}:
\begin{equation}
    r := \mathbb{E}_{p(\Omega)} [V(\hat{\pi}(\theta))] = \mathbb{E}_{p(\Omega)}\mathbb{E}_{x\sim p(x; \Omega)} \mathbb{E}_{\{x_i\}^N \sim p^N(x)} [U(x,\hat{\pi}(\theta))]
\end{equation}
where we have written the distributions $p(x;\Omega)$ as a function of the true parameters in our underlying model.\footnote{For example, in the GBM model, the true parameters are the growth rate $r$ and the volatility $\sigma$, and so $\Omega=(r, \sigma)$, and $p(\Omega) =p(r,\sigma)$.} In the main text, we have effectively assumed that $p(\Omega)$ is a Dirac delta distribution, but, in the more general case, it is possible that the true parameter is not known or cannot be accurately estimated, and it makes sense to assign a prior distribution to them. One can then find the optimal data augmentation with respect to $r$: $\theta^* = \arg\max_\theta r$.

\begin{table}[t!]
    \centering
    \begin{tabular}{c|c}
    \hline
    Financial Terms & Statistical Terms \\
    \hline
        Utility $U$ &  Loss function $L$ \\
        Expected Utility $V$ &  Risk $R$  \\
        Data Augmentation Parameter $\theta$ & Estimator $\hat{\theta}$ \\
        True Parameter $\Omega$ & True Parameter $\theta$ \\
        Prior $p(\Omega)$ & Prior $p(\theta)$ \\
    \end{tabular}
    \caption{Correspondence table between the concepts in our general theory and the classical statistical decision theory.}
    \label{tab:correspondence table}
\end{table}

See table~\ref{tab:correspondence table} for the list of correspondences. This analogy breaks down at the following point: the Bayesian estimator tries to find $\hat{\theta}$ that is as close to $\theta$ as possible, while, in our formulation, the goal is not to make data augmentation $\theta$ as close as possible to $\Omega$.

One might also hope to establish an analogous ``minimax" theory for the portfolio construction problem. This can be done simply by replacing the expectation over $p(\Omega)$ with a minimization over $\Omega$:
\begin{equation}
    r_{minimax} := \min_\Omega V(\hat{\pi}(\theta))
\end{equation}
and the best augmentation parameter can be found as the maximizer of this risk: $\theta^* = \max_\theta \min_{\Omega} V $.

\subsection{Proof for no data augmentation}\label{app: no augmentation proof}

when there is no data augmentation, $\mathbb{E}_{t} \left[  G_t(\pi) \right] = b_t r_t$ and $\text{Var}_{t} \left[ G_t(\pi) \right] = 0$.
One immediately see that the utility is then maximized at
\begin{equation}\label{eq: no augmentation strategy}
    \pi_t^* = \begin{cases}
        1, & \text{if } r_t\geq 0\\
        -1, & \text{if } r_t<0.
    \end{cases}
\end{equation}

We restate the theorem here.
\begin{proposition}
{\rm (Utility of no-data augmentation strategy.)} Let the strategy be as specified in Eq.~\eqref{eq: no augmentation strategy}, and let the price trajectory be generated with GBM in Eq.~\eqref{eq: geometric brownian motion} with initial price $S_0$, then the true utility is
\begin{equation}
    U = [1 - 2\Phi({-r}/{\sigma})]r - \frac{\lambda}{2} \sigma^2
\end{equation}
where $U(\pi)$ is the utility function defined in Eq.~\eqref{eq: portfolio objective}.
\end{proposition}
\textit{Proof.} 
For a time-dependent strategy $\pi^*_t$, the true utility is defined as\footnote{While we mainly use $\Theta(x)$ as the Heaviside step function, we overload this notation a little. When we write $\Theta(x>0)$, $\Theta$ is defined as the indicator function. We think that this is harmless because the difference is clearly shown by the argument to the function.}
\begin{equation}
    U(\pi^*) = \mathbb{E}_{S_0', S_1',...,S_T', S_{T+1}'}\left[\frac{1}{T}\sum_{t=1}^{T+1} \pi_t^* r_t' - \left(\frac{\lambda}{2T}\sum_{t=1}^T (\pi_t^* r_t')^2  -  \mathbb{E}_{S_0', S_1',...,S_T', S_{T+1}'}[\pi_t^* r_t']^2 \right)  \right]
\end{equation}
where $S_1',...,S_T',S_{T+1}'$ is an independently sampled distribution for testing, and $r_t':= \frac{S_{t+1}' -S_t'}{S_t'}$ are their respective returns. Now, we note that we can write the price update equation (the GBM model) in terms of the returns:
\begin{equation}
    S_{t+1} = (1 +r) S_t + \sigma_t S_t \eta_t \to r_t = r + \sigma \eta_t
\end{equation}
which means that $r_t \sim \mathcal{N}(r, \sigma^2)$ obeys a Gaussian distribution. Therefore,
\begin{equation}\label{eq: true utility proof}
    U(\pi^*) = \frac{r}{T}\sum_{t=1}^{T}\pi_t^* - 
    \frac{\lambda \sigma^2}{2T} \sum_{t=1}^{T} (\pi_t^*)^2. 
\end{equation}

Now we would like to average over $\pi^*_t$, because we also want to average over the realizations of the training set to make the true utility independent of the sampling of the training set (see Section~\ref{sec: general theory} for an explanation). 

Recall that the strategy is defined as 
\begin{equation}
    \pi_t^* = {\begin{cases}
        1, & \text{if } r_t\geq 0\\
        -1, & \text{if } r_t<0.
    \end{cases}} = \Theta(r_t \geq 1) - \Theta(r_t < 1)
\end{equation}
for a training set $\{S_0, ..., S_T\}$, and $\Theta$ is the Heaviside step function. 
We thus have that 
\begin{equation}
    \begin{cases}
        (\pi^*_t)^2 = 1;\\
        \mathbb{E}_{S_1,...S_{T+1}}[\pi_t^*] = \mathbb{E}_{S_1,...S_{T+1}}[ \Theta(r_t \geq 0) - \Theta(r_t < 0)] = 1 - 2\Phi({-r}/{\sigma})
    \end{cases}
\end{equation}
where $\Phi$ is the Gauss c.d.f. We can use this to average the utility over the training set; noticing that the training set and the test set are independent, we can obtain
\begin{align}
    U &= \mathbb{E}_{S_1,...S_{T+1}}[U(\pi^*)]\\
    &= \frac{1}{T}\sum_{t=1}^{T} [1 - 2\Phi({-r}/{\sigma})]r - \frac{\lambda}{2}\frac{1}{T}\sum_{t=1}^{T}\sigma^2  \\
    &= [1 - 2\Phi({-r}/{\sigma})]r - \frac{\lambda}{2} \sigma^2.
\end{align}
This finishes the proof. $\square$

\subsection{Proof for Additive Gaussian noise}\label{app sec: additive proof}
Before we prove the proposition, we first prove that the strategy is indeed the one given in Eq.~\eqref{eq: additive strategy}:
\begin{lemma}
    The maximizer of the utility function in Eq.~\ref{eq: train objective} with additive gaussian noise is
    \begin{equation}\label{eq: additive strategy}
        \pi^*_t(\rho) =\begin{cases}
        \frac{r_t S_t^2}{2 \lambda \rho^2} , &\text{if }  -1<\frac{r_t S_t^2}{2 \lambda \rho^2}<1; \\
        {\rm sgn} (r_t), &\text{otherwise.}
    \end{cases}  
    \end{equation}
\end{lemma}

\textit{Proof.} With additive Gaussian noise, we have
\begin{equation}
    \begin{cases}
        \mathbb{E}_{t} \left[ G_t(\pi) \right] = \pi_t \mathbb{E}_{t} \left[\tilde{r}_t \right] = \pi_t \mathbb{E}_{t} \left[\frac{S_{t+1} + \rho_{t+1} \epsilon_{t+1} - S_t -\rho_t \epsilon_t }{S_{t}}\right]  = \pi_t \frac{S_{t+1}-S_t}{S_t} = \pi_t r_t;\\
        \text{Var}_{t} \left[ G_t(\pi) \right] = \pi_t^2 \text{Var}_t[\tilde{r}_t] =\pi_t^2  \text{Var}_t\left[\frac{\rho_{t+1} \epsilon_{t+1} -\rho_t \epsilon_t}{S_t} \right] =  \frac{2\rho^2 \pi_t^2}{S_t^2}, \\
    \end{cases}
\end{equation}
where the last line follows from the definition for additive Gaussian noise that $\rho_1=...\rho_T =\rho$. The training objective becomes
\begin{align}
    \pi^*_t &= \arg\max_{\pi_t} \left\{ \frac{1}{T} \sum_{t=1}^{T} \mathbb{E}_{t} \left[G_t(\pi) \right] - \frac{\lambda}{2} \text{Var}_t[ G_t(\pi)] \right\}\\
    &= \arg\max_{\pi_t} \left\{ \frac{1}{T} \sum_{t=1}^{T} \pi_t r_t  - \lambda \frac{\rho^2 \pi_t^2}{S_t^2} \right\} .
\end{align}
This maximization problem can be maximized for every $t$ respectively. Taking derivative and set to $0$, we find the condition that $\pi_t^*$ satisfies
\begin{align}
    &\frac{\partial}{\partial \pi_t} \left(\pi_t r_t  - \lambda \frac{\rho^2 \pi_t^2}{S_t^2} \right) = 0\\
    \longrightarrow \quad & \pi^*_t(\rho) = \frac{r_t S_t^2}{2 \lambda \rho^2}.
\end{align}
By definition, we also have $|\pi_t|\leq 1$, and so
    \begin{equation}
        \pi^*_t(\rho) =\begin{cases}
        \frac{r_t S_t^2}{2 \lambda \rho^2} , &\text{if }  -1<\frac{r_t S_t^2}{2 \lambda \rho^2}<1; \\
        {\rm sgn} (r_t), &\text{otherwise,}
    \end{cases}  
    \end{equation}
which is the desired result. $\square$

We would like to comment that, although we paid special attention to enforcing the constraint $|\pi_t|\leq$ it is often not needed in practice because investors tend to be quite risk-averse, and it is hard to imagine that any investor would invest all his or her money in the financial market such that $\pi_t=1$. Mathematically, this means that it is often the case that $\lambda \geq \frac{|r_t|S_t^2}{2\lambda \rho^2}$. Therefore, for what comes, we assume that $\lambda \geq \frac{|r_t|S_t^2}{2\lambda \rho^2}$ for all $t$ for notational simplicity; note that, even without assumption, the conclusion that a multiplicative noise is the better kind of data augmentation will not change. Now we are ready to prove the proposition.

\begin{proposition}{\rm (Utility of additive Gaussian noise strategy.)} Let the strategy be as specified in Eq.~\eqref{eq: additive strategy}, and other conditions the same as in Proposition~\ref{prop: no augmentation}, then the true utility is
\begin{equation}
    U_{\rm Add} = \frac{r^2}{2\lambda \sigma^2 T} \mathbb{E}_{S_t} \left[\frac{(\sum_t r_t S_t)^2}{\sum_t (r_t S_t)^2}\Theta(\sum_t r_t S_t^2) \right].
\end{equation}
\end{proposition}

\textit{Proof.} The beginning of the proof is similar to the case with no data augmentation. Following the same procedure, we obtain an equation that is the same as Eq.~\eqref{eq: true utility proof}:
\begin{equation}
        U(\pi^*) = \frac{r}{T}\sum_{t=1}^{T}\pi_t^* - 
    \frac{\lambda \sigma^2}{2T} \sum_{t=1}^{T} (\pi_t^*)^2. 
\end{equation}
Plug in the preceding lemma, we have
\begin{equation}
     U(\pi^*) = \frac{r}{T}\sum_{t=1}^{T} \frac{r_t S_t^2}{2\lambda \rho^2} - 
    \frac{\lambda \sigma^2}{2T} \sum_{t=1}^{T}  \left(\frac{r_t S_t^2}{2\lambda \rho^2}\right)^2.
\end{equation}

This utility is a function of the data augmentation strength $\rho$. For a fixed training set, we would like to find the best $\rho$ that maximizes the above utility. Note that the maximizer of the utility is different depending on the sign of $\sum r_t S^2_t$. Taking derivative and set to $0$, we obtain that
\begin{equation}
    (\rho^*)^2 = \begin{cases}
        \frac{\lambda \sigma^2}{2r}\frac{\sum_t^T (r_t S_t^2)^2}{\sum_t^T r_t S_t^2}, &\text{if }\sum_t^T r_t S_t^2 > 0 \\
        \infty, &\text{otherwise.} \\
    \end{cases}
\end{equation}
Plug in to the previous lemma, we have
\begin{equation}
    \pi_t^*(\rho^*) =  \frac{r_t S_t^2}{2\lambda (\rho^*)^2} =\frac{ r r_t S_t^2}{ \sigma^2} \frac{\sum_t r_t S_t^2}{\sum_t (r_t S_t^2)^2} \Theta\left(\sum_t r_t S_t^2 \right).
\end{equation}
One thing to notice is that the optimal strength is independent of $\lambda$, which is an arbitrary value and dependent only on the investor's psychology. Plug into the utility function and take expectation with respect to the training set, we obtain
\begin{align}
    U_{\rm add} &= \mathbb{E}_{S_1,...,S_{T+1}} \left[ U(\pi^*(\rho^*))\right] \\
    & =  U(\pi^*) = \frac{r}{T}\sum_{t=1}^{T}\pi_t^*(\rho^*) - 
    \frac{\lambda \sigma^2}{2T} \sum_{t=1}^{T} [\pi_t^*(\rho^2)]^2\\
    &= \frac{r^2}{2\lambda \sigma^2 T} \mathbb{E}_{S_1,...,S_{T+1}} \left[\frac{(\sum_t r_t S_t)^2}{\sum_t (r_t S_t)^2}\Theta \left(\sum_t r_t S_t^2 \right) \right].
\end{align}
This finishes the proof. $\square$

\begin{remark}
    Notice that the term in the expectation generally depends on $T$ in a non-trivial way and cannot be obtained explicitly. However, it does not cause a problem since the final goal is to compare it with the result in the next section. 
\end{remark}

\subsection{Proof for General Multiplicative Gaussian noise}\label{app sec: general mult noise proof}
Before we prove the proposition, we first find the strategy for this case. Note that the term $\rho_t^2 + \rho_{t+1}^2$ appears repetitively in this section, and so we define a shorthand notation for it: $\frac{1}{2}(\rho_t^2 + \rho_{t+1}^2) := \gamma_t^2$.
\begin{lemma}\label{lemma: general multiplicative portfolio}
    The maximizer of the utility function in Eq.~\ref{eq: train objective} with multiplicative gaussian noise is
    \begin{equation}
        \pi^*_t(\rho) =\begin{cases}
        \frac{r_t S_t^2}{2 \lambda \gamma_t^2} =  \frac{r_t S_t^2}{\lambda (\rho_t^2 + \rho_{t+1}^2)}, &\text{if }  -1<\frac{r_t S_t^2}{2 \lambda \gamma_t^2}<1; \\
        {\rm sgn} (r_t), &\text{otherwise.}
    \end{cases}  
    \end{equation}
\end{lemma}

\textit{Proof.} With additive Gaussian noise, we have
\begin{equation}
    \begin{cases}
        \mathbb{E}_{t} \left[ G_t(\pi) \right] = \pi_t \mathbb{E}_{t} \left[\tilde{r}_t \right] = \pi_t \mathbb{E}_{t} \left[\frac{S_{t+1} + \rho_{t+1} \epsilon_{t+1} - S_t -\rho_t \epsilon_t }{S_{t}}\right]  = \pi_t \frac{S_{t+1}-S_t}{S_t} = \pi_t r_t;\\
        \text{Var}_{t} \left[ G_t(\pi) \right] = \pi_t^2 \text{Var}_t[\tilde{r}_t] =\pi_t^2  \text{Var}_t\left[\frac{\rho_{t+1} \epsilon_{t+1} -\rho_t \epsilon_t}{S_t} \right] =  \frac{(\rho_t^2 + \rho_{t+1})^2 \pi_t^2}{S_t^2}. \\
    \end{cases}
\end{equation}
We see that $\frac{1}{2}(\rho_t^2 + \rho_{t+1}^2):= \gamma_t^2$ replaces the role of $2\rho^2$ for additive Gaussian noise. The training objective becomes
\begin{align}
    \pi^*_t &= \arg\max_{\pi_t} \left\{ \frac{1}{T} \sum_{t=1}^{T} \mathbb{E}_{t} \left[G_t(\pi) \right] - \frac{\lambda}{2} \text{Var}_t[ G_t(\pi)] \right\}\\
    &= \arg\max_{\pi_t} \left\{ \frac{1}{T} \sum_{t=1}^{T} \pi_t r_t  - \lambda \frac{\gamma_t^2 \pi_t^2}{S_t^2} \right\} .
\end{align}
This maximization problem can be maximized for every $t$ respectively. Taking derivative and set to $0$, we find the condition that $\pi_t^*$ satisfies
\begin{align}
    &\frac{\partial}{\partial \pi_t} \left(\pi_t r_t  - \lambda \frac{\gamma_t^2 \pi_t^2}{S_t^2} \right) = 0\\
    \longrightarrow \quad & \pi^*_t(\gamma_t) = \frac{r_t S_t^2}{2 \lambda \gamma_t^2}.
\end{align}
By definition, we also have $|\pi_t|\leq 1$, and so
    \begin{equation}
        \pi^*_t(\rho) =\begin{cases}
        \frac{r_t S_t^2}{2 \lambda \gamma_t^2} , &\text{if }  -1<\frac{r_t S_t^2}{2 \lambda \gamma_t^2}<1; \\
        {\rm sgn} (r_t), &\text{otherwise,}
    \end{cases}  
    \end{equation}
which is the desired result. $\square$

\begin{remark}
    For fair comparison with the previous section, we also assume that $\lambda \geq \frac{|r_t|S^2}{\lambda \gamma_t^2}$. Again, this is the same as assume that the investors are reasonably risk-averse and is the correct assumption for all practical circumstances.
\end{remark}

\begin{proposition}{\rm (Utility of general multiplicative Gaussian noise strategy.)} Let the strategy be as specified in Eq.~\eqref{eq: additive strategy}, and other conditions the same as in Proposition~\ref{prop: no augmentation}, then the true utility is
\begin{equation}
    U_{\rm mult} = \frac{r^2}{2\lambda \sigma^2} [1-\Phi(-r/\sigma)].
\end{equation}
\end{proposition}

\textit{Proof.} Most of the proof is similar to the Gaussian case by replaing $\rho^2$ with $\gamma_t^2$. Following the same procedure, We have:
\begin{equation}
        U(\pi^*) = \frac{r}{T}\sum_{t=1}^{T}\pi_t^* - 
    \frac{\lambda \sigma^2}{2T} \sum_{t=1}^{T} (\pi_t^*)^2.
\end{equation}
Plug in the preceding lemma, we have
\begin{equation}
     U(\pi^*) = \frac{r}{T}\sum_{t=1}^{T} \frac{r_t S_t^2}{2\lambda \gamma_t^2} - 
    \frac{\lambda \sigma^2}{2T} \sum_{t=1}^{T}  \left(\frac{r_t S_t^2}{2\lambda \gamma_t^2}\right)^2.
\end{equation}

This utility is a function of of the data augmentation strength $\gamma_t$, and, unlikely the additive Gaussian case, can be maximized term by term for different $t$. For a fixed training set, we would like to find the best $\gamma_t$ that maximizes the above utility. Note that the maximizer of the utility is different depending on the sign of $r_t S^2_t$. Taking derivative and set to $0$, we obtain that
\begin{equation}
    (\gamma_t^*)^2 = \begin{cases}
        \frac{ \sigma^2}{2r} r_t S_t^2, &\text{if } r_t S_t^2 > 0 \\
        \infty, &\text{otherwise.} \\
    \end{cases}
\end{equation}
Plug in to the previous lemma, we have
\begin{equation}
    \pi_t^*(\gamma_t^*) =  \frac{r_t S_t^2}{2\lambda (\gamma_t^*)^2} =\frac{ r}{ \lambda \sigma^2} \Theta(r_t) .
\end{equation}
One thing to notice that the optimal strength is independent of $\lambda$, which is an arbitrary value and dependent only on the psychology of the investor. Plug into the utility function and take expectation with respect to the training set, we obtain
\begin{align}
    U_{\rm add} &= \mathbb{E}_{S_1,...,S_{T+1}} \left[ U(\pi^*(\rho^*))\right] \\
    &=  U(\pi^*) = \frac{r}{T}\sum_{t=1}^{T}\pi_t^*(\gamma_t^*) - 
    \frac{\lambda \sigma^2}{2T} \sum_{t=1}^{T} [\pi_t^*(\gamma_t^*)]^2 \\
    &= \frac{r^2}{\lambda \sigma^2} \mathbb[E]_t[\Theta(r_t)]  - \frac{r^2}{2\lambda \sigma^2}\Theta(r_t)\mathbb[E]_t[\Theta(r_t)] \\
    &= \frac{r^2}{2\lambda \sigma^2}[1-\Phi(-r/\sigma)]\\
    &=\frac{r^2}{2\lambda \sigma^2}\Phi(r/\sigma)
\end{align}
This finishes the proof. $\square$

This result can be directly compared to the results in the previous section, and the following remark shows that the multiplicative noise injection is the best kind of noise.

\begin{remark} {\rm (Infinite augmentation strength.)}
    For all of the theoretical results, there is a corner case when the optimal injection strength is equal to infinity, which leads to a non-investing portfolio $\pi =0$. This case requires special interpretation. This corner case is due to the fact the underlying model we use has a constant, positive expected price return equal to $r$, and so it leads to the bizarre data augmentation which effectively amounts to throwing away all the training points with $<0$ return. This is unnatural for a real market. It is possible for the real market to have short-term negative return when conditioned on the previous prices, and so one should not simply discard the negative points. Therefore, in our algorithm section, we recommend treating the training points with positive and negative return equally by taking the absolute value of the data augmentation strength and ignoring $\infty$ case.
\end{remark}

\subsection{Proof of Theorem~\ref{theo: remark}}\label{app sec: proof of remark}
\begin{remark}
Combining the above propositions, one can quickly obtain that, if $\sigma \neq 0$, then $U_{\rm mult} > U_{\rm add}$ and $U_{\rm mult} > U_{\rm no-aug}$ with probability $1$ (Proof in Appendix).
\end{remark}
\textit{Proof}. We first show that $U_{\rm mult} > U_{\rm add}$. Recall that 
\begin{align}
    U_{\rm Add} &= \frac{r^2}{2\lambda \sigma^2 T} \mathbb{E}_{S_t} \left[\frac{(\sum_t r_t S_t)^2}{\sum_t (r_t S_t)^2}\Theta\left(\sum_t r_t S_t^2 \right) \right] \\
    &\leq  \frac{r^2}{2\lambda \sigma^2 T} \mathbb{E}_{S_t} \left[\frac{\left(\sum_t r_t S_t \Theta(r_t > 0)\right)^2}{\sum_t (r_t S_t)^2}\Theta \left(\sum_t r_t S_t^2 \right) \right] \\
    &\leq \frac{r^2}{2\lambda \sigma^2 T} \mathbb{E}_{S_t} \left[\frac{(\sum_t r_t S_t \Theta(r_t > 0))^2}{\sum_t (r_t S_t)^2} \right]\\
    & \leq_{\text{(Cauchy Inequality)}} \frac{r^2}{2\lambda \sigma^2 T} \mathbb{E}_{S_t} \left[\sum _t \frac{ ( r_t S_t \Theta(r_t > 0))^2}{ (r_t S_t)^2} \right]\\
    & = \frac{r^2}{2\lambda \sigma^2 T} \mathbb{E}_{S_t} \left[\sum _t  \Theta(r_t > 0) \right] \\
    & = \frac{r^2}{2\lambda \sigma^2 T} \left[\sum _t   \mathbb{P}(r_t > 0) \right] \\
    & = \frac{r^2}{2\lambda \sigma^2}  \Phi(r/t)  = U_{\rm mult}.
\end{align}
The Cauchy equality holds if and only if $S_1 = ... =S_{T+1}$; this event has probability measure $0$, and so, with probability $1$, $U_{\rm add} < U_{\rm mult}$.

Now we prove the second inequality. Recall that 
\begin{equation}
    U_{\rm no-aug} =[1 - 2\Phi({-r}/{\sigma})]r - \frac{\lambda}{2} \sigma^2.
\end{equation}
We divide into $2$ subcases. Case $1$: $\lambda > \frac{r}{\sigma^2}$. We have
\begin{equation}
    U_{\rm no-aug} < - 2r \Phi({-r}/{\sigma}) < 0 < U_{\rm mult}.
\end{equation}

Case $2$: $0<\lambda \leq  \frac{r}{\sigma^2}$. We have
\begin{align}
    U_{\rm no-aug} &< [1 - 2\Phi({-r}/{\sigma})]r \\
    & < \Phi({r}/{\sigma}) r \\
    &\leq \frac{r^2}{2\lambda \sigma^2}  \Phi(r/t)  = U_{\rm mult}.
\end{align}
This finishes the proof. $\square$

\subsection{Augmentation for a naive multiplicative noise}\label{app sec: naive multiplicative}
In the discussion and experiment sections in the main text, we also mentioned a ``naive" version of the multiplicative noise. The motivation for this kind of noise is simple, since the underlying noise in the theoretical models are all of the form $\sigma^2 S_t^2$, and so it is tempting to also inject noise that mimicks the underlying noise. It turns out that this is not a good idea.

In this section, we let $\rho_t = \rho_0 S_t$ for some positive, time-independent $\rho_0$. Our goal is to find the optimal $\rho_0$. With the same calculation, one finds that the learned portfolio is given by the same formula in Lemma~\ref{lemma: general multiplicative portfolio}:
\begin{equation}\label{eq: naive strategy}
    \pi^*_t(\rho) = \frac{r_t S_t^2}{2\lambda \gamma_t^2}. 
\end{equation}
With this strategy, one can find the optimal noise injection strength to be given by the following proposition.
\begin{proposition}
    Let the portfolio be given by Eq.~\eqref{eq: naive strategy} and let the price be generated by the GBM, then the optimal noise strength is
    \begin{equation}\label{eq: naive strength}
        (\rho_0^*)^2 = \begin{cases}
            \frac{\sum_t^T r_t^2}{\sum_t^T r_t}, &\text{if } \sum_t^T r_t > 0 \\
            \infty, &\text{otherwise.} \\
        \end{cases}
    \end{equation}
\end{proposition}
\textit{Proof.} As before, We have:
\begin{equation}
        U(\pi^*) = \frac{r}{T}\sum_{t=1}^{T}\pi_t^* - 
    \frac{\lambda \sigma^2}{2T} \sum_{t=1}^{T} (\pi_t^*)^2.
\end{equation}
Plug in the portfolio, we have
\begin{equation}
     U(\pi^*) = \frac{r}{T}\sum_{t=1}^{T} \frac{r_t S_t^2}{2\lambda \gamma_t^2} - 
    \frac{\lambda \sigma^2}{2T} \sum_{t=1}^{T}  \left(\frac{r_t S_t^2}{2\lambda \gamma_t^2}\right)^2.
\end{equation}

Plug in $\gamma_t^2 = \rho_0^2 S_t^2$ and take derivative, we obtain that
\begin{equation}
    (\rho_0^*)^2 = \begin{cases}
        \frac{\sum_t^T r_t^2}{\sum_t^T r_t}, &\text{if } \sum_t^T r_t > 0 \\
        \infty, &\text{otherwise.} \\
    \end{cases}
\end{equation}
This finishes the proof. $\square$

\subsection{Data Augmentation for a Stationary Portfolio}\label{app sec: stationary portfolio}
While the main theory focused on the case with a dynamic portfolio that is updated through time, we also present a study of the stationary portfolio in this section. While this kind of portfolio is less relevant for deep-learning-based finance, we study this case to show that, even in this setting, there is still strong motivation to inject noise of strength $r_t S_t^2$. Now we state the formal definition of a stationary porfolio.

\begin{definition}
    A portfolio $\{\pi_t\}_{t=1}^T$ is said to be stationary if $\pi_t = \pi$ for some constant $\pi$ for all $t$.
\end{definition}
In the language of machine learning, this corresponds to choosing our model as having only a single parameter, whose output is input independent:
\begin{equation}
    f(x) = \pi.
\end{equation}

In traditional finance theory, stationary portfolios have been very important. In practice, most portfolios are ``approximately" stationary, since most portfolio managers tend to not to change their portfolio weight at a very short time-scale unless the market is very unstable due to market failure or external information injection. For a stationary portfolio, one still would like to maximize the utility function given in Eq.~\ref{eq: train objective}.

For conciseness, we only examine the case when we inject a general time-dependent noise. The curious readers are encouraged to examine the cases with no data augmentation and with constant data augmentation. As before, the following lemma gives the portfolio of the empirical utility. Again, we use the shorthand: $\frac{1}{2}(\rho_t^2 + \rho_{t+1}^2) := \gamma_t^2$. For illustrative purpose, we have ignore the corner cases of $\pi$ being greater than $1$ or smaller than $-1$.
\begin{lemma}
    The stationary portfolio that maximizes the utility function in Eq.~\ref{eq: train objective} with multiplicative gaussian noise is
    \begin{equation}\label{eq: optimal stationary portfolio}
        \pi^*_t(\rho) = \frac{\sum_t^T r_t}{2\lambda \sum_t^T (\gamma_t^2/ S_t^2)}
    \end{equation}
\end{lemma}

\textit{Proof sketch}. The proof follows almost the same as the previous sections. With a slight difference that $\pi$ is no more time-dependent and can be taken out of the sum.

\begin{proposition}\label{prop: stationary optimal augmentation strength}
    Let the portfolio be given in Eq.~\eqref{eq: optimal stationary portfolio}, then an augmentation strength satisfying the relation
    \begin{equation}
        (\gamma_t^*)^2 = c r_t S_t^2
    \end{equation}
    is an optimal data augmentation for constant $c = \frac{2\sigma^2}{r}$.
\end{proposition}

\textit{Proof Sketch}. The proof is also simple and very similar to the proofs for the dynamic portfolio case. One first solve for the optimal augmentation strength $\gamma_t^*$ and find that it satisfies the following relation
\begin{equation}
    \sum_t \frac{(\gamma^*_t)^2}{S_t^2} = c \sum_t^T r_t
\end{equation}
with $c= \frac{2\sigma^2}{r}$, and then it suffices to check that the following is one solution
\begin{equation}
    (\gamma^*_t)^2 = cr_t S_t^2.
\end{equation}
This finishes the proof sketch. $\square$

The curious readers are encouraged to check the intermediate steps. We see that, even for a stationary portfolio case, there is still strong motivation for using a augmentation with strength proportional to $r_t S_t^2$. 

We would like to compare this with the best achievable stationary portfolio, which is solved by the following proposition.
\begin{proposition}
    \textrm{(Optimal Stationary Portfolio)}. The optimal stationary portfolio for GBM is $\pi^*_{stat} = \frac{r}{\lambda \sigma^2}$, i.e., for any other portfolio $\pi$, $U(\pi^*_{stat}) \geq U(\pi)$.
\end{proposition}
\textit{Proof.} It suffices to find the maximizer portfolio of the true utility:
\begin{equation}
    \pi^*_{stat} = \arg \max_{\pi} \left\{\pi r - \frac{\lambda}{2}\pi^2 \sigma^2 \right\}.
\end{equation}
The solution is simple and given by
\begin{equation}
    \pi^{*}_{stat} = \frac{r}{\lambda \sigma^2}.
\end{equation}
This completes the proof. $\square$

Note that the above optimality result holds for both in-sample counterfactual utility and out-of-sample utility. This proposition can be seen as the discrete-time version of the famous Merton's portfolio solution \citep{merton1969lifetime}, where the optimal stationary portfolio is also found to be $\frac{r}{\lambda \sigma^2}$. In fact, it is well-known that, for a static market, the stationary portfolios are optimal, but this is beyond the scope of this work \citep{CoverInformationTheory}.

Combining the above two propositions, one obtain the following theorem.
\begin{theorem}
    The stationary portfolio obtained by training with data augmentation strength in given in Proposition~\ref{prop: stationary optimal augmentation strength} is optimal, i.e., it is no worse than any other stationary portfolio.
\end{theorem}
\textit{Proof.} Plug in $\gamma = c r_t S_t^2$, we have that the trained portfolio is $\pi^* = \frac{r}{\lambda \sigma^2}$, which is equivalent to the optimal stationary portfolio, and we are done. $\square$ 

This shows that, even for a stationary portfolio, it is useful to use the proposed data augmentation technique.

\end{document}